\documentclass[11pt]{article}

\usepackage[preprint]{acl}

\usepackage{times}
\usepackage{latexsym}

\usepackage[T1]{fontenc}

\usepackage[utf8]{inputenc}

\usepackage{microtype}

\usepackage{inconsolata}

\usepackage{graphicx}

\usepackage{tcolorbox}
\usepackage{amsmath}
\usepackage{amssymb}
\usepackage{mathtools}
\usepackage{bbm}
\usepackage{amsthm}
\usepackage{amsfonts}

\usepackage[table]{xcolor} 
\definecolor{rowgreen}{rgb}{0.9, 1, 0.9} 
\definecolor{rowblue}{rgb}{0.9, 0.95, 1} 
\definecolor{rowred}{rgb}{1, 0.9, 0.9}   
\usepackage{multirow}

\usepackage{listings}
\usepackage{xcolor}
\usepackage[textsize=tiny]{todonotes}

\usepackage{microtype}
\usepackage{graphicx}
\usepackage{subcaption}
\usepackage{booktabs} 
\usepackage{siunitx}
\usepackage{algorithm}
\usepackage{algorithmic}
\title{Continuity and Ordinality Matter: Constraining Time Series Tokens \\for Effective Time Series Analysis with Large Language Models}

\author{
Musheng Li, Ziying Zhang, Cheng Jin, Yuantao Gu\thanks{Corresponding author.} \\
Department of Electronic Engineering \\
Tsinghua University \\
{\small \texttt{\{lms21, ziying-z21, jinc21\}@mails.tsinghua.edu.cn}, 
 \texttt{gyt@tsinghua.edu.cn}}
}

\begin{document}
\maketitle
\begin{abstract}
Token-based time series large language models (TS-LLMs) have emerged as a promising direction for time series analysis and reasoning. However, prior studies largely overlook the inherent continuity and ordinality of time series tokens, which substantially limits model performance. In this paper, we argue that preserving these properties in time series token embeddings is crucial for the effectiveness of token-based TS-LLMs. To this end, we propose \textbf{COM} (\textit{Continuity and Ordinality Matter}), a continuity- and ordinality-aware strategy that integrates geometric constraints into both the initialization and training stages. Empirical results on multiple time series analysis benchmarks demonstrate that COM consistently improves the performance of token-based TS-LLMs, achieving competitive results and strong generalizability. Code is available at \url{https://anonymous.4open.science/r/COM}.
\end{abstract}

\section{Introduction}

Time series analysis plays a crucial role in a wide range of real-world applications, including finance, healthcare, traffic, energy, and industrial monitoring~\cite{sezer2020financial-finance, pinto2021ensemble-energy, li2015trend-traffic, haoyietal-informer-2021-ett, hewage2020temporal-weather}. Despite its importance, conventional time series modeling is often highly task-specific: different tasks, datasets, and domains typically require separately designed architectures or independently trained models~\cite{liu2024todynet, Yuqietal-2023-PatchTST, Zeng2022AreTE-Dlinear,ramaswamy2000efficient-anomaly}, which substantially limits their scalability and general applicability. Inspired by the recent success of large language models (LLMs) in unified sequence modeling and general-purpose reasoning, time series large language models (TS-LLMs) have emerged as a promising direction for building more generalizable models for time series analysis~\cite{xie2025chatts, chattime, jin2023time-llm, chow2024towards-chow}.

Existing TS-LLMs generally fall into four paradigms: text-, vision-, alignment-, and token-based approaches. Text-based methods~\cite{kong2025time-time-mqa} regard time series as plain text. Vision-based methods~\cite{zhuang2024see-tama} visualize time series and leverage vision-language models (VLMs)~\cite{bai2025qwen2.5-vl}. Alignment-based methods~\cite{xie2025chatts} introduce a time series encoder and align it with the LLM backbone. Token-based methods~\cite{chattime,quinlan2026chat-ts,ansari2024chronos} quantize numerical values into a finite set of time series tokens and extend the LLM vocabulary accordingly. Among these paradigms, token-based methods are particularly promising for building general-purpose TS-LLMs, as they provide a unified token space for time series and text, scale naturally with long-context LLMs, and largely retain the pretrained capabilities of LLM backbones.

However, existing token-based TS-LLMs overlook the intrinsic continuity and ordinality of time series in both token embedding initialization and subsequent training~\cite{chattime}. Specifically, newly introduced time series token embeddings are typically initialized with random Gaussian vectors, while the training process lacks explicit constraints to preserve the continuous and ordinal structure among time series tokens. This may result in a fragmented embedding space and distorted time series representations, ultimately limiting the model performance.



\begin{figure*}[ht]
    \centering
    \includegraphics[width=\textwidth]{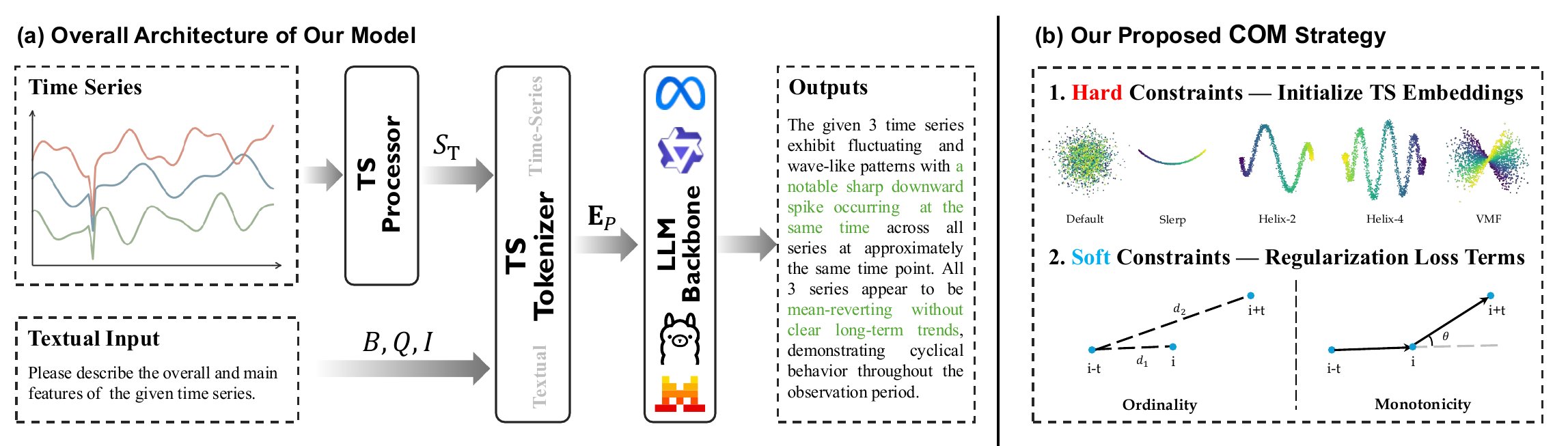}
    \caption{
    Overall illustration of our work. 
    \textbf{(a)} Our approach follows the token-based TS-LLM paradigm, consisting of a TS-Processor, a TS-Tokenizer, and an LLM backbone. 
    \textbf{(b)} To preserve the continuity and ordinality of TS tokens, we introduce the \textbf{COM} (\textit{Continuity and Ordinality Matter}) strategy, which combines \emph{hard constraints} through embedding initialization with \emph{soft constraints} through dedicated regularization loss terms.
    }
    \label{fig:overview_new}
\end{figure*}

To bridge this gap, we propose \textbf{COM} \textit{(Continuity and Ordinality Matter)}, a continuity- and ordinality-aware strategy for token-based TS-LLMs. COM explicitly preserves the structural properties of time series tokens at two stages: during initialization, it injects structured priors through controllable TS token embedding initialization; during training, it regularizes the embedding geometry with dedicated loss terms. We further design quantitative metrics to evaluate continuity and ordinality in the embedding space. Extensive ablation studies on time series question-answering and reasoning benchmarks verify the effectiveness of these constraints, and additional evaluations on other tasks demonstrate the broader potential of COM.

Empirical results demonstrate that COM consistently improves the performance ceiling and convergence efficiency of TS-LLMs. The performance gains are strongly correlated with the geometric properties of TS token embeddings, highlighting the importance of continuity and ordinality preservation in the embedding space. Across diverse model families and parameter scales, COM maintains robust effectiveness, indicating strong generalizability. Moreover, evaluations on three TSA tasks, namely question-answering, classification, and description, show that COM-enhanced TS-LLMs achieve competitive performance against task-specific baselines, demonstrating their potential for general-purpose TSA.

Our contributions are summarized as follows:
\begin{enumerate}
    \item We identify the lack of continuity and ordinality modeling as a critical limitation of existing token-based TS-LLMs, and propose COM, a strategy that explicitly preserves these structural properties at both the initialization and training stages.
    \item Through comprehensive comparative and ablation studies, we show that COM consistently improves the performance and convergence efficiency of TS-LLMs. We further reveal that such improvements are closely correlated with the continuity and ordinality of TS embeddings, offering insights into the mechanism and design of effective constraints.
    \item We achieve competitive performance against specialized models across diverse TSA tasks, demonstrating the generalizability, robustness, and universal potential of COM-enhanced TS-LLMs.
\end{enumerate}

\section{Preliminary}
\label{sec:preliminary}

In the field of time series large language models (TS-LLMs), diverse time series analysis (TSA) tasks can be unified under a time series question-answering (TSQA) framework. Formally, denote a dataset as $\mathcal{D} = \{(T_i, Q_i, G_i), \mathcal{M}\}_{i=1}^{N}$, where $\mathcal M$ represents the task-specific evaluation metric. For the $i$-th sample, $T_i = \{\mathbf{t}_{i,1}, \dots, \mathbf{t}_{i,C}\}$ comprises $C$ channels of time series data, where $\mathbf{t}_{i,j} \in \mathbb{R}^{L_j}$ denotes the time series in the $j$-th channel with an arbitrary length $L_j$. 

The question $Q_i$ is formulated based on the specific TSA task requirements. For instance, in \text{multiple-choice QA}, $Q_i$ directs the model to select the most plausible option from a given candidate set; 
in \text{classification}, $Q_i$ instructs the model to categorize the sequence into a class $\mathcal{S}_c$. 
Accordingly, $G_i$ serves as the ground truth aligned with the $Q_i$.

Given a TS-LLM denoted by $\Phi$ and its generating function $f_{\Phi}(\cdot)$, the inference process is defined as:
\begin{equation}
    A_i = f_{\Phi} \ (T_i, Q_i, P_i)
\end{equation}
where $A_i$ is the generated response by TS-LLM $\Phi$, and $P_i$ represents the prompt template encompassing task descriptions, context, and operational constraints. The overall performance of $\Phi$ on dataset $\mathcal{D}$ is evaluated as:
\begin{equation}
\label{eq:perf}
    \text{Perf.}(\Phi, \mathcal{D}) = \frac{1}{N} \sum_{i=1}^{N} \mathcal{M}(G_i, A_i)
\end{equation}
In practice, the evaluation metric $\mathcal{M}$ is task-dependent and can be instantiated as accuracy, F1-score, BLEU, ROUGE-L, semantic similarity, MAE, MSE, or other appropriate metrics.

\section{Methodology}
\subsection{Token-Based TS-LLM Architecture}
\label{sec:architecture}

The architecture of our model, as illustrated in Figure~\ref{fig:overview_new} (a), follows a {token-based TS-LLM} paradigm. It consists of three core components: a {TS-Processor}, a {TS-Tokenizer}, and a {pre-trained LLM backbone}.

\textbf{TS Token Vocabulary and Processing.} 
To bridge the gap between continuous numerical values and discrete linguistic tokens, we expand the LLM's vocabulary with $N$ time series tokens, denoted as $\mathcal{V}_{ts}$. Given a precision $\epsilon$, the number of TS tokens is $N = \lceil 2/\epsilon \rceil + 1$.
The \text{TS-Processor} transforms the raw time series $T$ into a token sequence $S_T$ through three stages:
(1) \textbf{Normalization}: The entire time series $T$ is scaled by its global maximum absolute value: $\hat{\mathbf{t}}_j = \frac{\mathbf{t}_j}{\max_{\mathbf{t} \in T} (|\mathbf{t}|)}$,  mapping all values into the range $[-1, +1]$. 
(2) \textbf{Quantization}: The normalized values are mapped to the nearest TS tokens in $\mathcal{V}_{ts}$ via nearest-neighbor quantization. 
(3) \textbf{Prompt Construction}: We extract statistical metadata (e.g., min, max, mean, std) and integrate them with the quantized time series using a pre-defined template $\mathcal{T}$. This process is formalized as:
\begin{equation}
\label{eq:ts_processor}
    S_T = \text{TS-Processor}(T; \epsilon, \mathcal{T})
\end{equation}
As shown in Figure~\ref{fig:template_example}, our method inherently supports multivariate inputs with different sequence lengths.

\textbf{Prompt and Tokenization.} 
The generated $S_T$ is then wrapped into a comprehensive prompt $P$ using a task-specific wrapper $\mathcal{W}(\cdot)$:
\begin{equation}
    \label{eq:get_prompt}
    P = \mathcal{W}(S_T, B, Q, I)
\end{equation}
where $B$ denotes the background information of the task, $Q$ is the specific question, and $I$ represents operational instruction constraints (e.g., output format or option restrictions), or reasoning requirements (e.g., "think step-by-step"). 

The \text{TS-Tokenizer} processes $P$ into a sequence of token IDs, and then maps them into embeddings:
\begin{equation}
    \mathbf{E}_P = \text{TS-Tokenizer}(P)
\end{equation}
where $\mathbf{E}_P \in \mathbb{R}^{L \times D}$ represents the embedding tensor of the sequence $P$, which consists of textual and TS tokens.

\textbf{LLM Backbone and Inference.} 
The embeddings $\mathbf{E}_P$ are then fed into the LLM backbone $\Phi$. The model generates the answer $A$ in an autoregressive manner:
\begin{equation}
    A = f_{\Phi}(\mathbf{E}_P)
\end{equation}

By incorporating TS tokens directly into the vocabulary, $\Phi$ can generate sequences that interleave natural language with time series segments. This unified manner enables a single model to handle diverse downstream tasks.

\subsection{COM: Continuity- and Ordinality-Aware TS Token Modeling}
\label{subsec:com}

Existing token-based TS-LLMs typically treat TS tokens as ordinary discrete vocabulary entries~\cite{chattime}. However, they overlook a fundamental modality discrepancy: unlike linguistic tokens (e.g., ``apple'', ``play''), which lack an inherent notion of order, time series tokens possess intrinsic continuity and ordinality. For instance, \textit{``-0.1, 0.0, 0.1, 0.2''} follows a rigorous increasing order. We argue that enforcing these properties within the embedding space is a prerequisite for effective TSA and superior performance. Interestingly, this argument is further supported by the visualization of TS embeddings in ChatTime~\cite{chattime}: although no explicit geometric constraints are imposed, the learned embeddings already exhibit emerging continuous and ordered structures, as shown in Figure~\ref{fig:chattime_visualization}. Consequently, we propose \textbf{COM} \textit{(Continuity and Ordinality Matter)}, a continuity- and ordinality-aware strategy for token-based TS-LLMs.




\textbf{Initialization Stage: Manifold Initialization.}
To instantiate the continuity and ordinality of TS embeddings $\mathbf{E}_{ts}$, we initialize them as a continuous, ordered manifold rather than via random initialization. This process is governed by two priors:
(1) \textbf{Distribution Consistency:} newly added TS embeddings should match the mean and variance of existing embeddings, which ensures the LLM backbone identifies TS inputs within a familiar numerical range~\cite{dobler2025awedist-new-token,chung2024stable-embed}.
(2) \textbf{Low-dimensional Manifold Hypothesis:} since TS tokens represent scalar values, which contain significantly lower information than semantic text, they should reside in a low-dimensional subspace. This is also consistent with recent studies showing that features of image and text modalities tend to concentrate on low-dimensional manifolds~\cite{song2019generative-langevin, niu2021low-dim, wikipedia_manifold_hypothesis}.
Based on these priors, we design a series of initialization schemes
 (see Appendix~\ref{app:init_implement} for details). Figure~\ref{fig:visualization} illustrates the 2D PCA visualization of TS embeddings related to different initialization schemes.

\begin{figure}[h]
  \centering

  \begin{subfigure}[t]{0.32\linewidth}
    \centering
    \includegraphics[width=\linewidth]{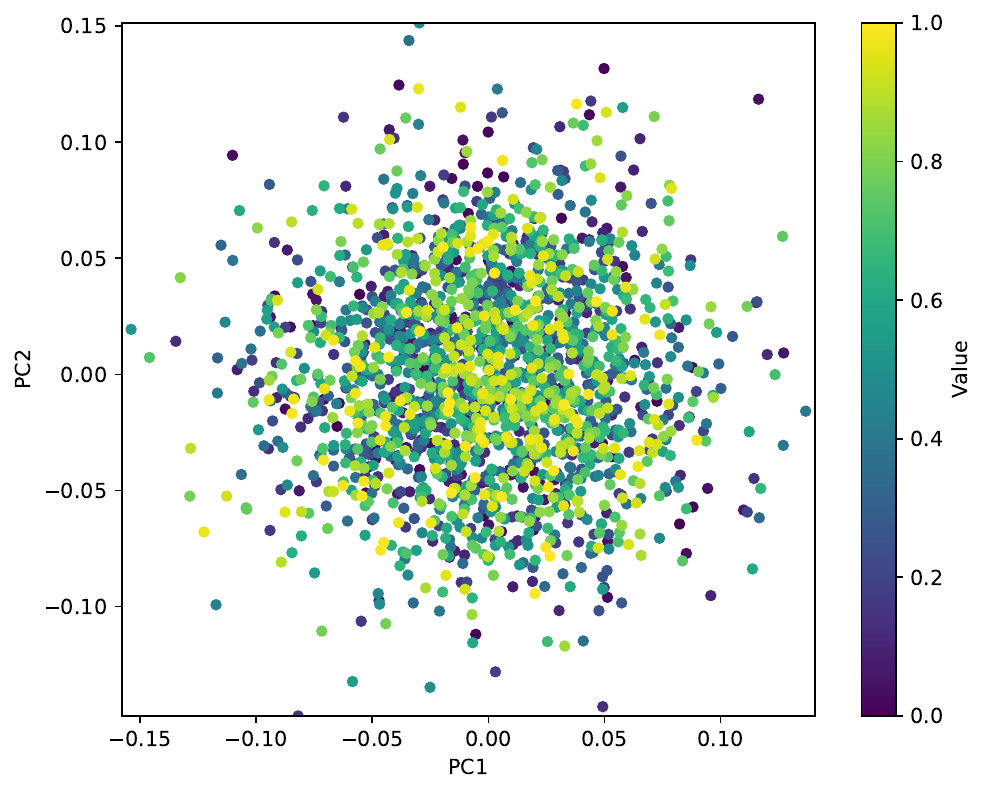}
    \caption{Default}
  \end{subfigure}
  \hfill
  \begin{subfigure}[t]{0.32\linewidth}
    \centering
    \includegraphics[width=\linewidth]{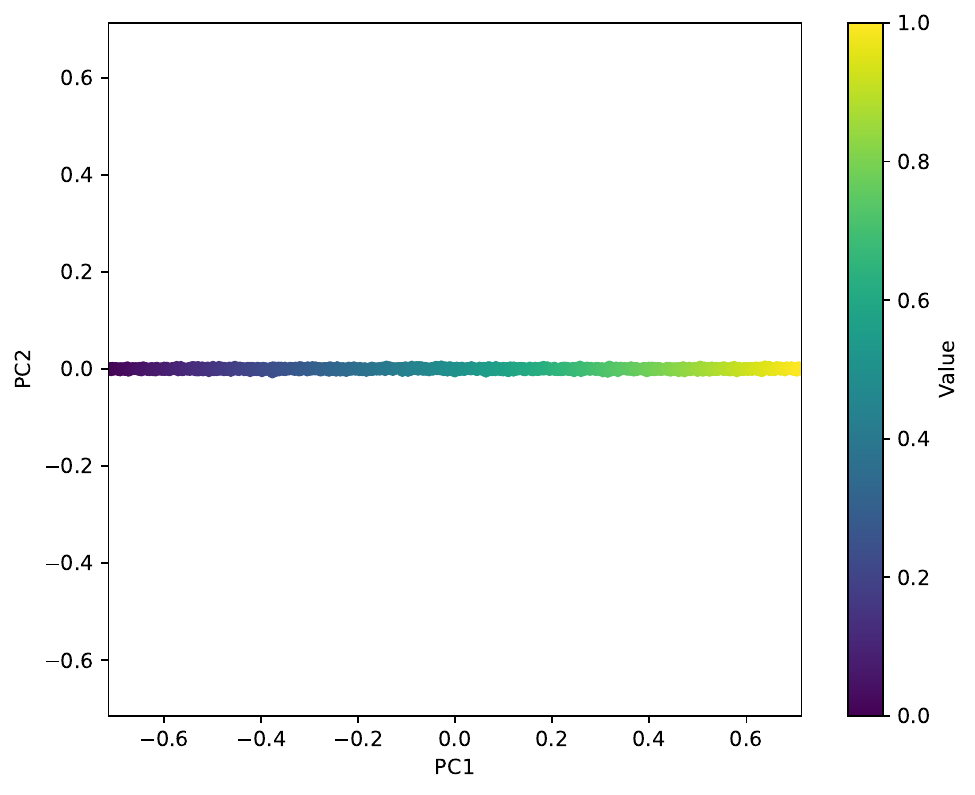}
    \caption{PCA-Main}
  \end{subfigure}
  \hfill
  \begin{subfigure}[t]{0.32\linewidth}
    \centering
    \includegraphics[width=\linewidth]{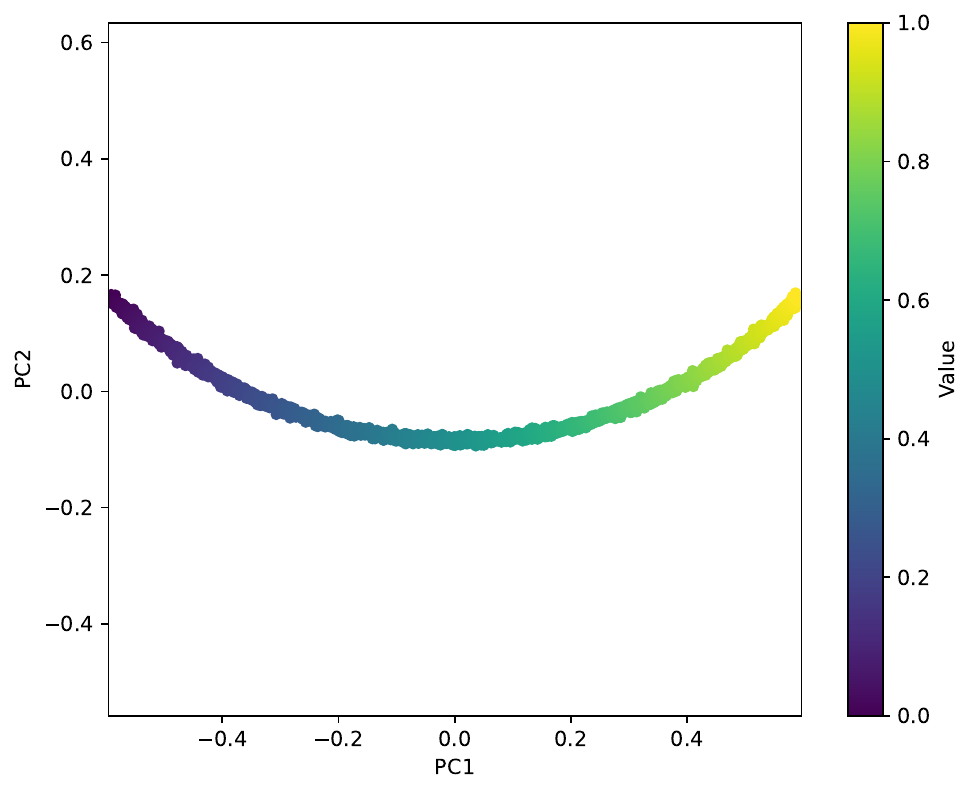}
    \caption{Slerp}
  \end{subfigure}

  \vspace{0.1em}

  \begin{subfigure}[t]{0.32\linewidth}
    \centering
    \includegraphics[width=\linewidth]{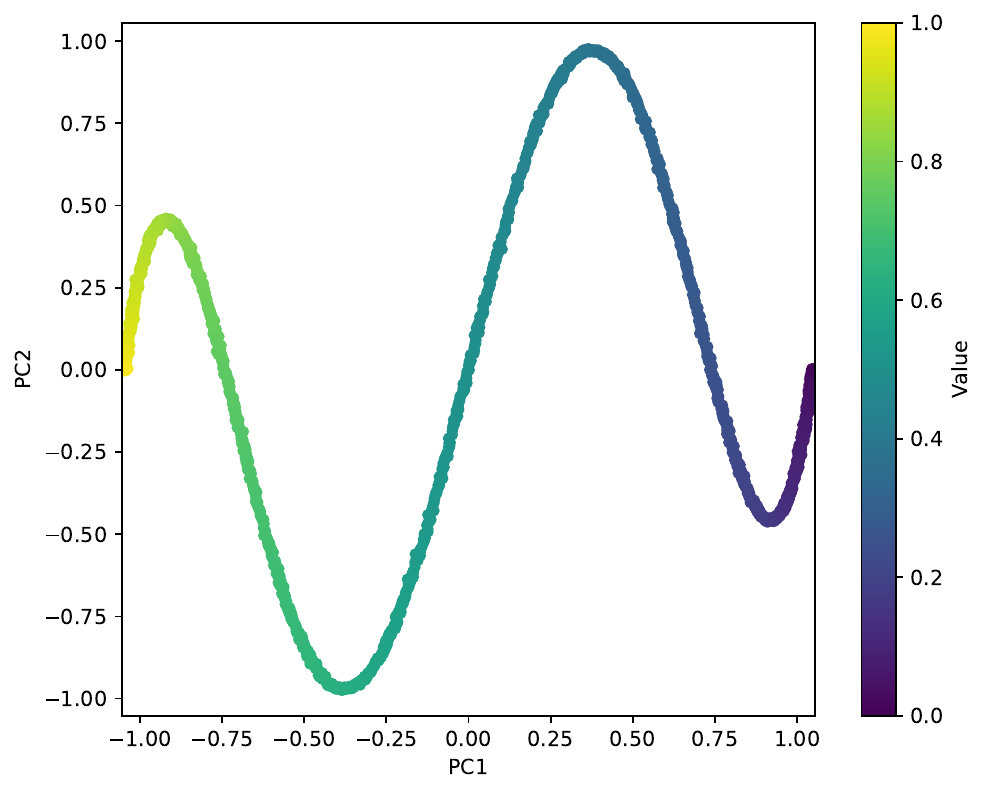}
    \caption{Helix-2}
  \end{subfigure}
  \hfill
  \begin{subfigure}[t]{0.32\linewidth}
    \centering
    \includegraphics[width=\linewidth]{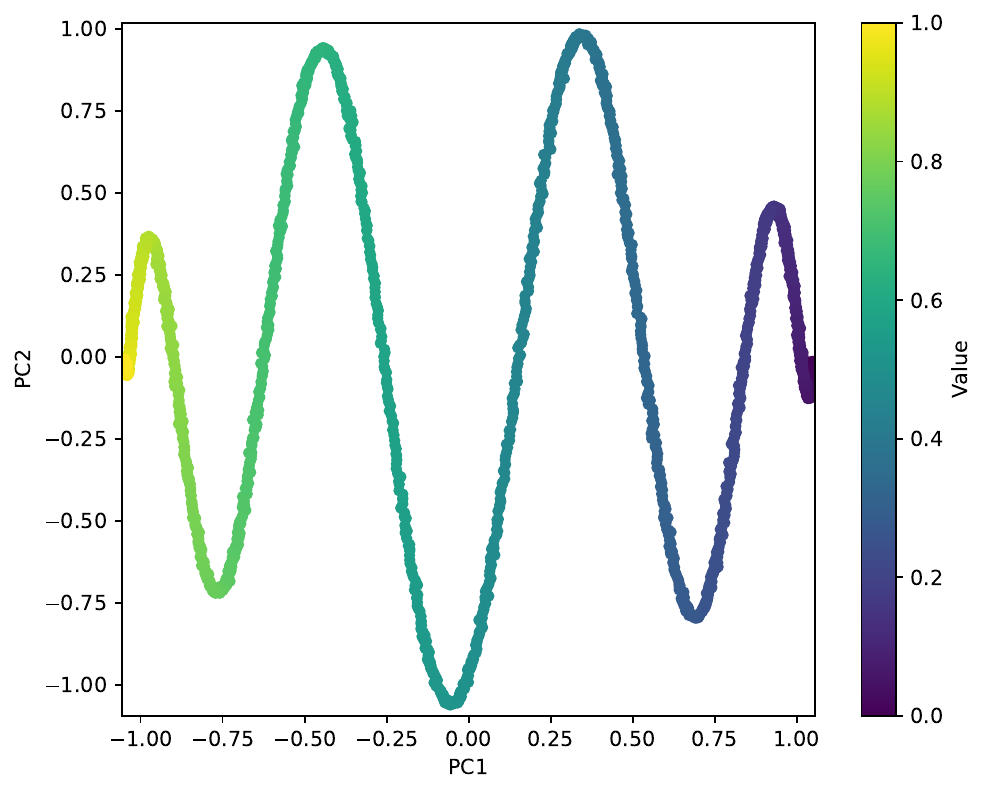}
    \caption{Helix-4}
  \end{subfigure}
  \hfill
  \begin{subfigure}[t]{0.32\linewidth}
    \centering
    \includegraphics[width=\linewidth]{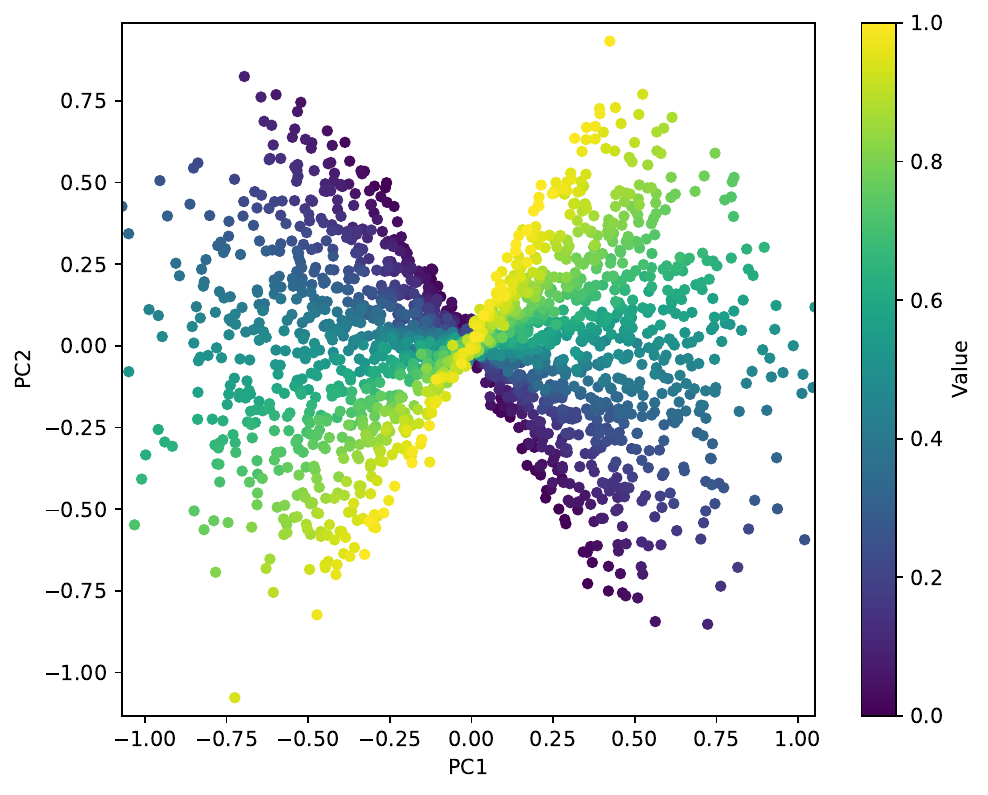}
    \caption{VMF}
  \end{subfigure}

  \caption{2D PCA visualizations of TS embeddings under different initialization schemes (hard constraints). The color denotes TS values ranging from $-1$ to $+1$. Details are provided in Appendix~\ref{app:init_implement}.}
  \label{fig:visualization}
\end{figure}

\textbf{Training Stage: Geometry-Preserving Regularization.}
To dynamically constrain the manifold's geometric properties during training, we introduce two regularization terms:

\textit{1) Ordinality Loss ($\mathcal{L}_{\text{ord}}$).}
To preserve ordinal consistency, we encourage the embedding-space distances among TS tokens to reflect their numerical ordering.
For a step size $k$, $\mathbf{e}'_{i-k}$ should be closer to $\mathbf{e}'_i$ than $\mathbf{e}'_{i-2k}$:
\begin{equation}
\resizebox{0.999\columnwidth}{!}{$
    \mathcal{L}_{\text{ord}} = \frac{1}{N-2k} \sum_{i=2k}^{N} \max(0, \|\mathbf{e}'_i - \mathbf{e}'_{i-k}\|_2 - \|\mathbf{e}'_i - \mathbf{e}'_{i-2k}\|_2 - m_{\text{ord}})
$}
\end{equation}
To mitigate the ``curse of dimensionality'' and preserve meaningful distance semantics, we project the embeddings into a 3D PCA subspace using $\mathbf{P} \in \mathbb{R}^{D \times 3}$, yielding $\mathbf{e}'_i = \mathbf{P}^\top \mathbf{e}_i$.

\textit{2) Monotonicity Loss ($\mathcal{L}_{\text{mono}}$).} To promote smooth embedding trajectories and suppress erratic fluctuations such as zig-zag patterns, we encourage consecutive displacement vectors in the embedding space to be directionally aligned:
\begin{equation}
\resizebox{0.999\columnwidth}{!}{$
    \mathcal{L}_{\text{mono}} = \frac{1}{N-2k} \sum_{i=k}^{N-k} \max(0, -\text{cos\_sim}(\mathbf{e}'_i - \mathbf{e}'_{i-k}, \mathbf{e}'_{i+k} - \mathbf{e}'_i) - m_{\text{mono}})
$}
\end{equation}
where $m_{\text{ord}}$ and $m_{\text{mono}}$ denote controllable margin thresholds in the ordinality and monotonicity losses, respectively. By adjusting the step $k$, we achieve multi-granularity constraints: a small $k$ constrains local structure, while a large $k$ constrains global structure. We also derive metrics $\mathcal{R}_{\text{ord}}$ and $\mathcal{R}_{\text{mono}}$ by setting margins to $0$ to evaluate geometric properties. The overall process is summarized in Algorithm~\ref{alg:com}.

\subsection{Training Objective and Pipeline}
\label{subsec:training}

\textbf{Optimization Objective.} 
Following the standard supervised fine-tuning (SFT) paradigm, we optimize the TS-LLM backbone $\Phi$ (together with hard-constrained TS embedding weights $\varphi_{h}$) by minimizing the negative log-likelihood of the ground truth answer $G$, conditioned on the composite prompt $P$ (Equation~\ref{eq:get_prompt}). Let $G$ be a sequence of tokens $\{g_1, g_2, \dots, g_M\}$. The cross-entropy loss is formulated as:
\begin{equation}
    \mathcal{L}_{\text{CE}} = - \frac{1}{M} \sum_{t=1}^{M} \log P_{\Phi, \varphi_h}(g_t \mid P, g_{<t})
\end{equation}
where $g_{<t}$ denotes the preceding tokens in the answer. To enforce the geometric constraints (Sec.~\ref{subsec:com}), we incorporate the soft regularization terms into the final objective:
\begin{equation}
    \mathcal{L}_{\text{total}} = \mathcal{L}_{\text{CE}} + \lambda_{\text{ord}}\mathcal{L}_{\text{ord}} + \lambda_{\text{mono}}\mathcal{L}_{\text{mono}}
\end{equation}
where $\lambda_{\text{ord}}$ and $\lambda_{\text{mono}}$ are hyperparameters controlling the strength of the ordinality and monotonicity losses.

\begin{figure*}[ht!]
    \centering
    \begin{subfigure}[t]{0.30\textwidth}
        \centering
        \includegraphics[width=\textwidth]{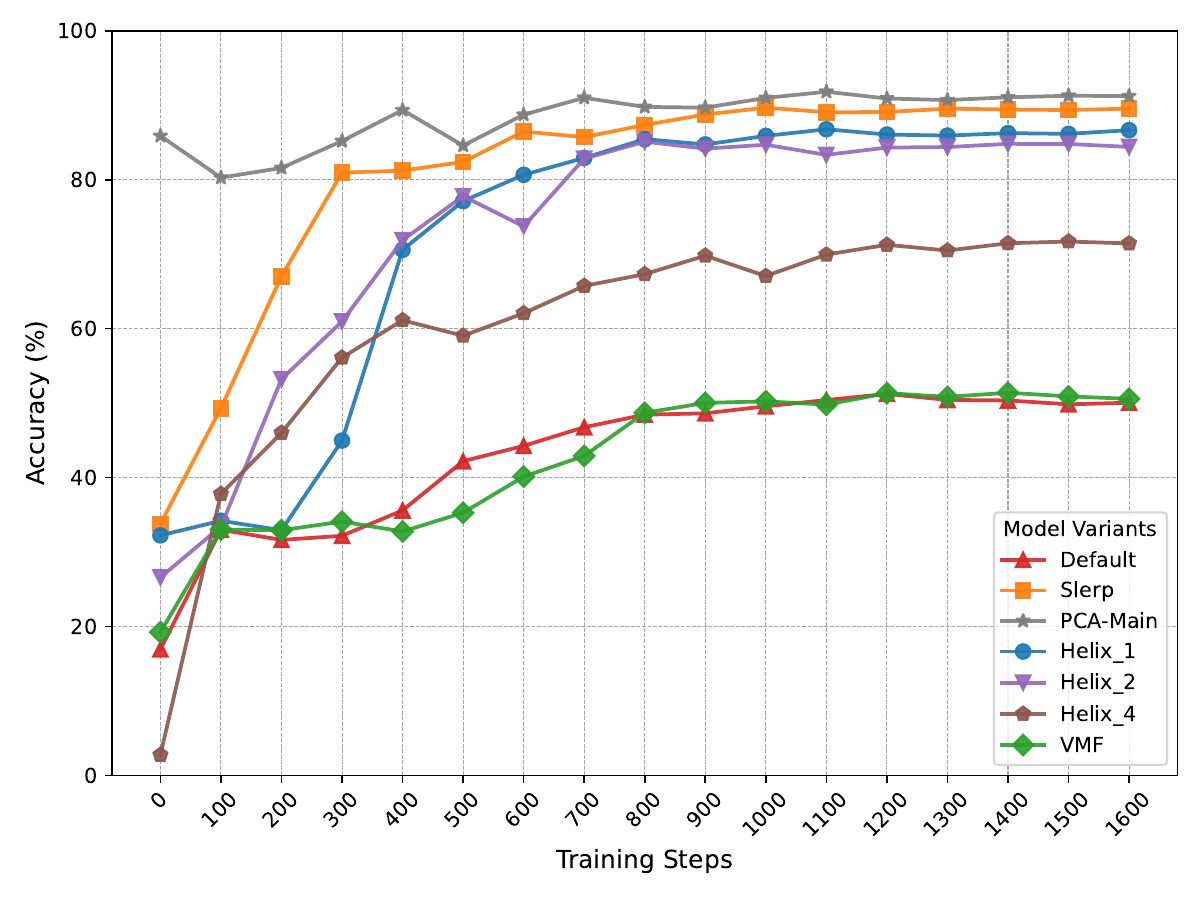}
        \caption{Acc. vs. steps on different variants}
        \label{fig:init-different}
    \end{subfigure}
    \hfill
    \begin{subfigure}[t]{0.30\textwidth}
        \centering
        \includegraphics[width=\textwidth]{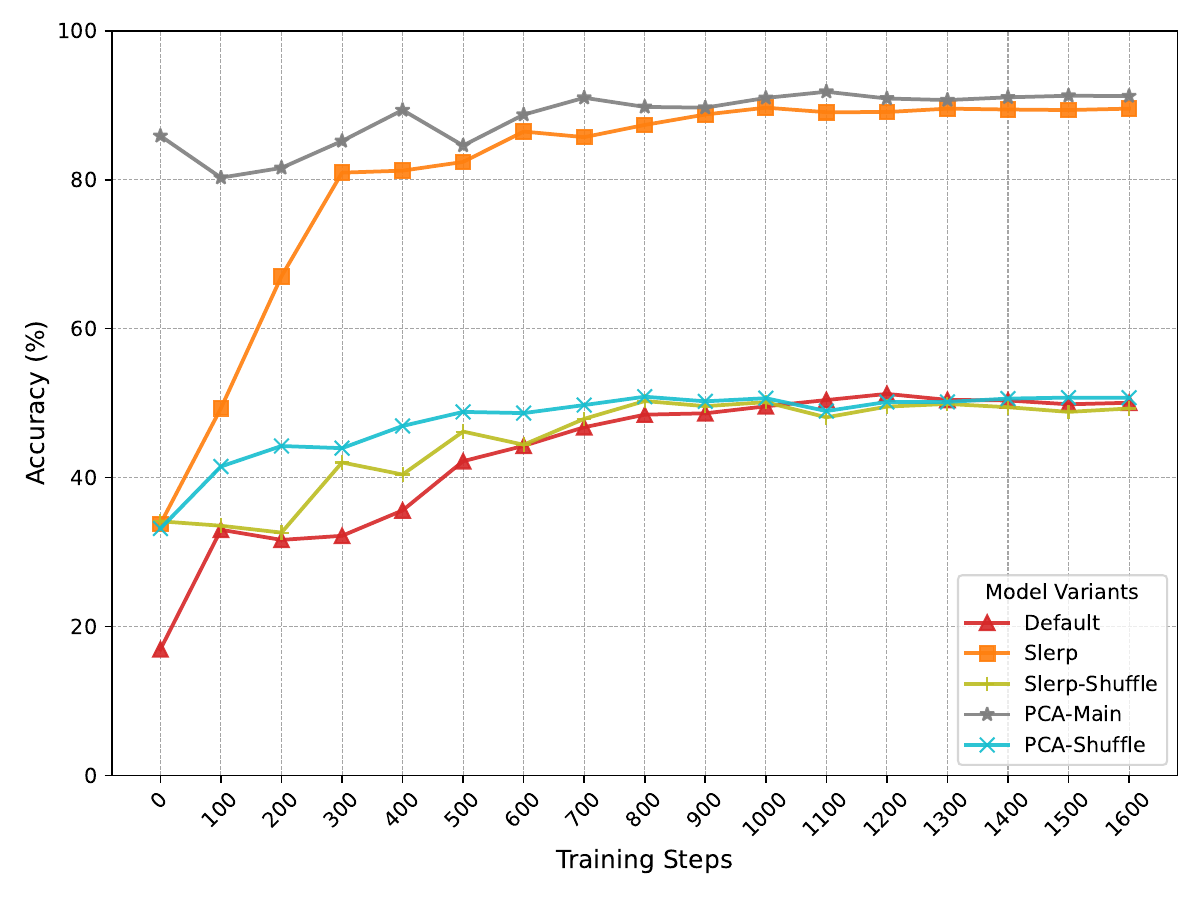}
        \caption{Ablation on shuffled variants}
        \label{fig:init-shuffle}
    \end{subfigure}
    \hfill
    \begin{subfigure}[t]{0.34\textwidth}
        \centering
        \includegraphics[width=\textwidth]{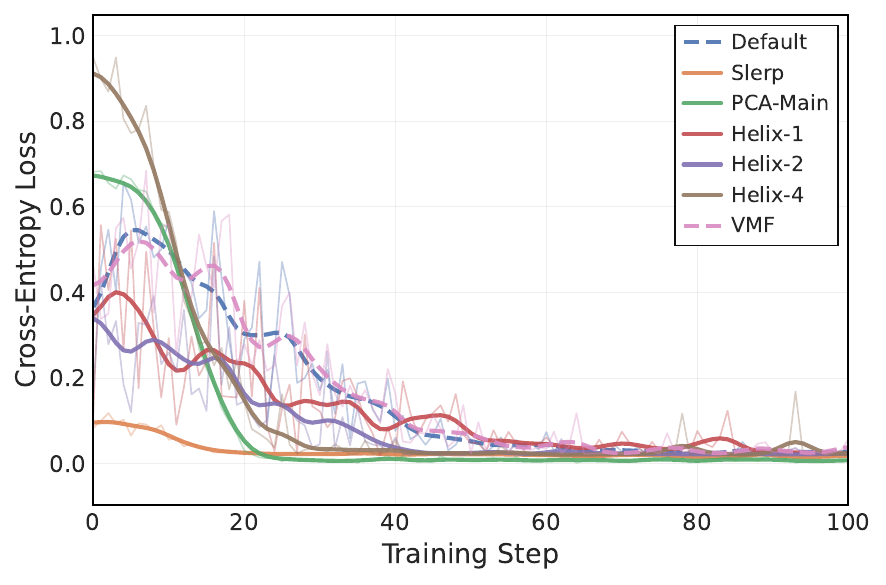}
        \caption{Training loss of different variants}
        \label{fig:ce-loss}
    \end{subfigure}

    \caption{
    The figure illustrates the accuracy trajectories of different initialized model variants (descriptions provided in Appendix~\ref{app:init_implement}) over training steps, together with the evolution of cross-entropy loss during training.
    }
    \label{fig:init-efficacy}
\end{figure*}

\textbf{Training Pipeline.} 
The pipeline consists of two phases:

\textit{1) Self-Distillation Alignment.} 
General-purpose LLMs already possess zero-shot capabilities for TSA.
A common practice is to distill this knowledge by training smaller models on synthetic data generated by teacher models. However, significant discrepancies in linguistic styles between different model families can introduce distributional shifts, causing the model to overfit to linguistic artifacts rather than learning key TSA patterns.
To mitigate this, we propose a \text{self-distillation alignment}, utilizing larger models from the \text{same family} to synthesize instruction-tuning data. 
Experiments show this phase effectively activates the model's preliminary TSA capabilities (Appendix~\ref{app:efficacy_self_alignment} ).

\textit{2) Downstream Fine-Tuning.} 
The aligned model serves as a task-agnostic foundation for subsequent adaptation. In this phase, we perform SFT on varying datasets to enable the model for downstream tasks. 
\section{Experiments}

\label{sec:experiments}

We conduct a comprehensive evaluation of COM-enhanced token-based TS-LLMs. Our experiments are designed to systematically address the following four research questions (RQs):

{\textbf{RQ1 (Efficacy of Hard Constraints):} Does the manifold initialization (hard constraints) improve performance consistently, and what is the underlying explanation?}

{\textbf{RQ2 (Mechanism of Soft Constraints):} How do the regularization losses (soft constraints) affect different models, what is the relationship between gains and regularization? What are the best practices for controlling margin and granularity?}

{\textbf{RQ3 (Generalizability):} As a lightweight strategy, can COM be seamlessly integrated into various LLMs to provide consistent performance gains?}

{\textbf{RQ4 (Universal Potential):} Can our model demonstrate its potential as a universal TSA model by achieving competitive performance across diverse fundamental tasks?}

See Appendix~\ref{app:exp_setup} for experimental settings.

\subsection{RQ1: Efficacy of Hard Constraints}
\label{subsec:exp_hard_constraints}

We systematically investigate the efficacy of manifold initialization (hard constraints) in terms of TS-LLMs' time series modeling capabilities (RQ1) on TSQA~\cite{chattime} dataset.

\paragraph{Impact of Geometric Priors.}
We benchmark five initialization schemes representing distinct geometric priors: \textit{Slerp}, \textit{PCA-Main}, \textit{Helix} (with varying turns $k$), and \textit{VMF}, utilizing Gaussian random initialization (\textit{Default}) as the baseline. 

As illustrated in Figure~\ref{fig:init-different}, the training trajectories reveal a distinct stratification correlated with geometric attributes: \textbf{Top Tier} (PCA-Main, Slerp, Helix-1/2) benefits from high continuity and strict ordinality, achieving superior accuracy. \textbf{Middle Tier} (Helix-4) exhibits degradation due to increased structural complexity, where excessive spiral turns induce aliasing between tokens of adjacent cycles. Conversely, \textbf{Bottom Tier} (Default, VMF), characterized by discontinuity or numerical overlap, fails to effectively capture TS dependencies, resulting in the worst performance. Notably, PCA-Main achieves competitive performance even without training, and ordered initialization variants consistently outperform disordered counterparts in the untrained setting, indicating that the hard constraints could provide a favorable starting point for adaptation. This finding remains consistent on the MMTS-InWild dataset (Table~\ref{tab:inwild_init}).

\paragraph{The Critical Role of Ordinality.}
To investigate the contribution of spatial ordinality, we conduct an ablation experiment by introducing a \textit{Shuffle} operation to the Slerp and PCA-Main schemes. This operation shuffles the TS token indices while preserving the overall geometric shape of the manifold. 
As shown in Figure~\ref{fig:init-shuffle}, the shuffled variants suffer a catastrophic performance drop, reverting to the level of the Default baseline. This confirms that geometric shape alone is insufficient; the \text{ordinality} of TS tokens along the manifold is decisive for modeling time series. Furthermore, comparing the convergence rates in Figure~\ref{fig:ce-loss}, ordered models (solid lines) demonstrate significantly faster convergence compared to disordered counterparts (dashed lines), validating proper hard constraints effectively ease the optimization.

\begin{table}[ht]
\caption{Accuracy (\%) on TSQA dataset across different initialization variants. $^*$ denotes shuffled models. T, S, V, and O represent Trend, Seasonality, Volatility, and Outliers, respectively. Background colors indicate \colorbox{rowgreen}{Top}, \colorbox{rowblue}{Middle}, and \colorbox{rowred}{Bottom} tiers.}
\label{tab:tsqa_init}
\vskip 0.15in
\begin{center}
\begin{small}
\begin{sc}
\setlength{\tabcolsep}{5pt}
\resizebox{\columnwidth}{!}{
\begin{tabular}{l|c|cccc}
\toprule
\textbf{Variants} & \textbf{Avg}. & \textbf{T}. & \textbf{S}. & \textbf{V}. & \textbf{O}.  \\
\midrule
\rowcolor{rowgreen} PCA-Main     & \textbf{91.23} & 98.77 & 68.39 & 98.46 & 99.22  \\
\rowcolor{rowgreen} Slerp        & \underline{89.54} & 98.69 & 66.64 & 95.72 & 97.04  \\
\rowcolor{rowgreen} Helix1       & 86.65 & 97.21 & 67.14 & 84.48 & 98.09 \\
\rowcolor{rowgreen} Helix2       & 84.40 & 98.11 & 67.06 & 85.77 & 86.43  \\
\rowcolor{rowblue} Helix4        & 71.44 & 95.15 & 62.04 & 56.35 & 72.35 \\
\rowcolor{rowred} PCA-Main$^*$   & 50.71 & 76.50 & 39.38 & 39.94 & 46.78 \\
\rowcolor{rowred} VMF            & 50.56 & 73.21 & 31.27 & 44.46 & 53.22 \\
\rowcolor{rowred} Default        & 50.04 & 76.83 & 36.45 & 41.79 & 44.70 \\
\rowcolor{rowred} Slerp$^*$      & 49.27 & 76.25 & 34.87 & 38.72 & 47.04 \\
\bottomrule
\end{tabular}
}
\end{sc}
\end{small}
\end{center}
\vskip -0.1in
\end{table}

Moreover, Table~\ref{tab:tsqa_init} reveals a performance difference driven by task sensitivity. \textbf{Shape-dominant tasks} (Trend, Outliers), which focus on global patterns or obvious anomalies, prove relatively robust to embedding granularity, yielding nontrivial performance even under misordered initialization. Conversely, \textbf{Value-sensitive tasks} (Seasonality, Volatility) demand precise modeling of numerical magnitudes and periodic shifts; in this regime, models with continuous and ordered embeddings significantly outperform disordered counterparts, underscoring the necessity of manifold initialization for capturing subtle time series features.

\vspace{0.5em}
\noindent\fbox{
\parbox{0.98\linewidth}{
\textbf{Finding 1:} Hard constraints are decisive for TSA: \textit{ordinality} preserves relative ordinal relationships, while \textit{continuity} ensures smooth transitions across adjacent tokens. These constraints optimize the performance, accelerate convergence, and ensure precise numerical modeling.
}
}

\subsection{RQ2: Mechanism of Soft Constraints}
\label{subset:exp_soft_constraints}

\begin{figure}[h]
    \centering
    \includegraphics[width=\linewidth]{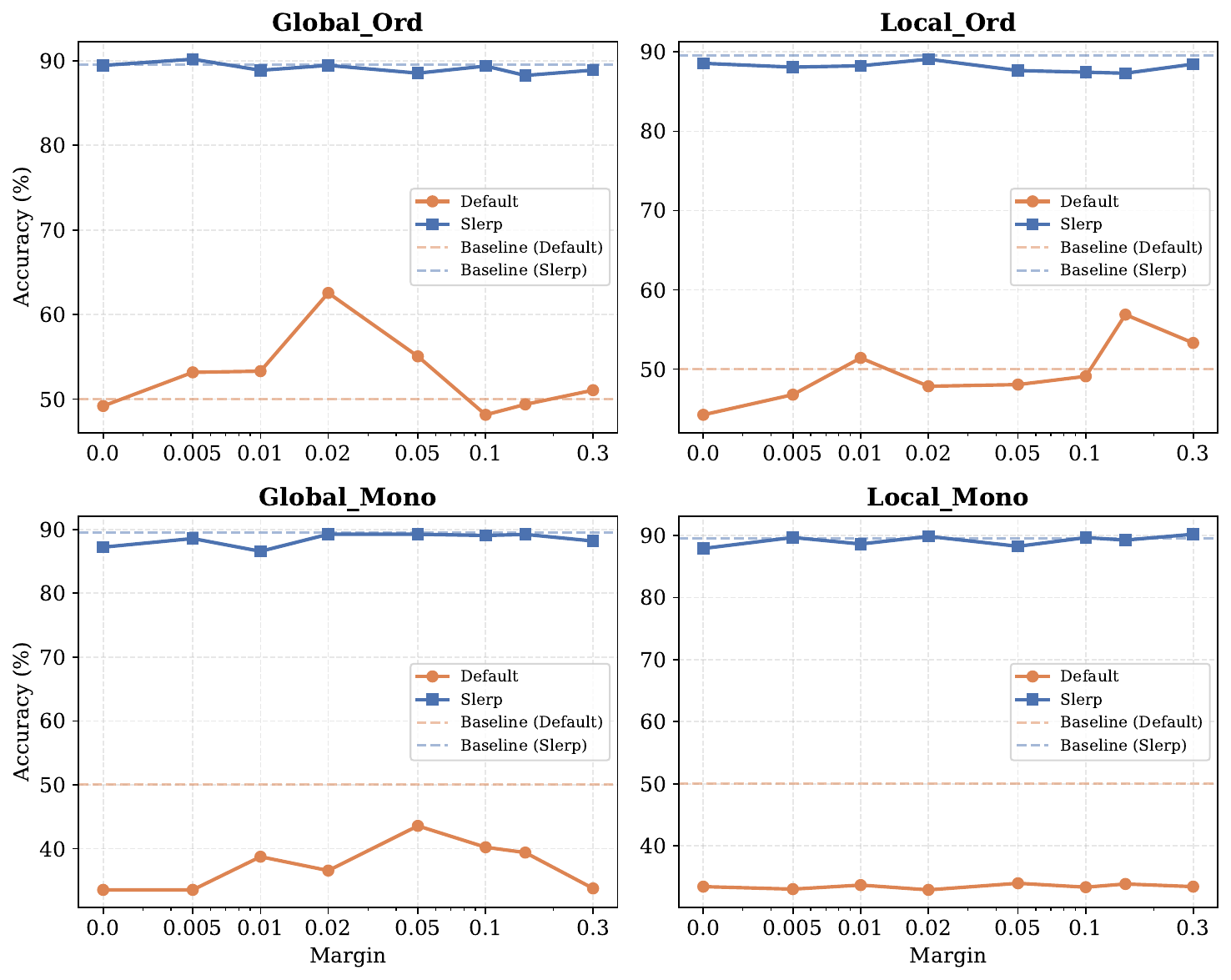}
    \caption{Ablation study of four regularization loss variants on the TSQA dataset. \texttt{Ord} and \texttt{Mono} denote ordinality and monotonicity losses, while \texttt{Local} and \texttt{Global} refer to 1 and 100 steps, respectively. The dashed lines represent the baseline performance without soft constraints.}
    \label{fig:ablation}
\end{figure}

To investigate the underlying mechanisms and optimal practices of soft constraints, we conduct a systematic ablation study of the regularization losses on TSQA dataset. As formulated previously, the regularization consists of \textit{Ordinality Loss} ($\mathcal{L}_{\text{ord}}$) and \textit{Monotonicity Loss} ($\mathcal{L}_{\text{mono}}$). By configuring the steps $s \in \{1, 100\}$, we derive four loss variants: $\mathcal{L}_{{\{\text{ord, mono}\}}\times{\{\text{local, global}\}}}$, targeting local and global geometric characteristics, respectively. We evaluate these constraints on two representative model variants: \text{Default} and \text{Slerp}. Experimental results with varying margins are illustrated in Figure~\ref{fig:ablation}.

The results yield several key insights. For the \textbf{Default} model, which inherently exhibits high structural disorder, the introduction of global and local ordinality losses significantly boosts performance (up to $\sim 12\%$). Conversely, the monotonicity loss leads to a performance degradation below the baseline. This phenomenon can be attributed to the substantial loss gap stemming from random initialization. The high initial monotonicity penalty likely weakens the effective gradient contribution from the cross-entropy objective, thereby hurting optimization. This observation highlights the sensitivity of the monotonicity loss and suggests that realizing its full benefit may require more carefully calibrated training recipes. In contrast, for the \textbf{Slerp} model with favorable geometric properties, the four regularization terms have a marginal impact, with performance fluctuating slightly around the baseline.

\begin{figure}[h]
    \centering
    \includegraphics[width=\linewidth]{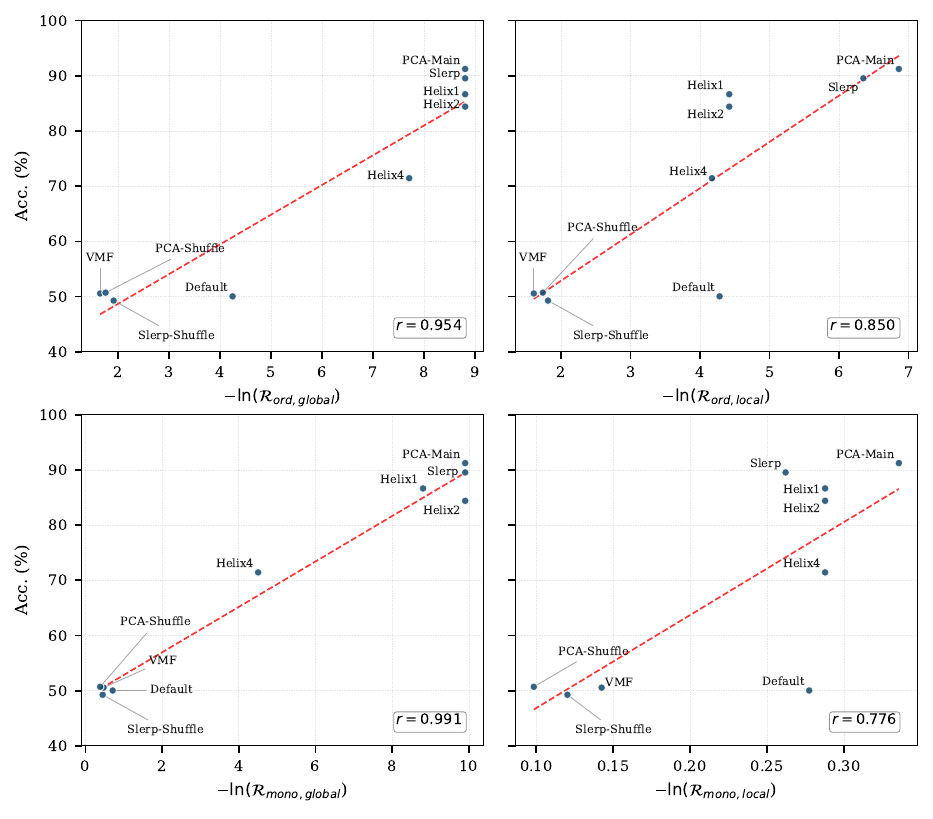}
    \caption{Linear regression analysis between model performance and the negative logarithm of metrics $\mathcal{R}_{\{\text{ord, mono}\} \times \{\text{local, global}\}}$. The consistent positive correlation suggests lower $\mathcal{R}$ (stronger geometric regularity) is predictive of better downstream performance.}
    \label{fig:regression}
\end{figure}

To further quantify the relationship between geomteric attributes and model performance, we design four metrics ($\mathcal{R}_{{\{\text{ord, mono}\}}\times{\{\text{local, global}\}}}$) by setting the margin thresholds to $0$. These metrics reflect the extent of ordinality and monotonicity in TS embedding space, where higher values indicate poorer geometric consistency. We perform a linear regression between the model performance and the negative logarithm of these metrics. As shown in Figure~\ref{fig:regression}, the performance exhibits a strong positive correlation with the global metrics (Pearson's $|r| > 0.9$) and a positive trend with the local metrics. This correlation validates the regularization losses as effective proxies for geometric properties and demonstrates a strong alignment with model performance. While other factors (e.g., training epochs, geometric shapes) also affect the performance, this correlation remains consistent and aligns with theoretical expectations.

\vspace{0.5em}
\noindent\fbox{
\parbox{0.98\linewidth}{
\textbf{Finding 2:} {Appropriately selected soft constraints $\mathcal{L}$ significantly enhance model performance. The metrics $\mathcal{R}$ serve as reliable proxies for geometric regularity, exhibiting a robust correlation with task performance.}
}}

\subsection{RQ3: Generalizability of COM}
\label{sec:generalizability}

While the efficacy of the COM strategy has been established on the \text{Qwen2.5-3B-Instruct}~\cite{qwen2025qwen25technicalreport-qwen2.5} backbone in previous sections,
it is imperative to verify its generalizability across different backbones. To this end, we extend our evaluation to LLM backbones of different families and parameter scales, including \text{Llama-3.2-3B-Instruct}~\cite{grattafiori2024llama-llama3}, \text{Llama-3.2-1B-Instruct},  \text{Gemma-2-2b-it}~\cite{team2024gemma-gemma2}, and \text{Qwen2.5-7B-Instruct}~\cite{qwen2025qwen25technicalreport-qwen2.5}, while maintaining previous experimental configuration. To streamline the analysis, we focus on comparing the model performance under \textbf{Default} and \textbf{Slerp} initialization schemes.

\begin{table}[t]
\centering
\caption{Accuracy (\%) on the TSQA dataset across different LLM backbone families and parameter scales.}
\label{tab:generalizability}
\resizebox{\columnwidth}{!}{
\begin{tabular}{@{}lcccccccccc@{}}
\toprule
\multirow{2}{*}{\textbf{Models}} 
& \multicolumn{2}{c}{\textbf{Llama-1B}} 
& \multicolumn{2}{c}{\textbf{Gemma-2B}} 
& \multicolumn{2}{c}{\textbf{Qwen-3B}} 
& \multicolumn{2}{c}{\textbf{Llama-3B}} 
& \multicolumn{2}{c}{\textbf{Qwen-7B}} \\
\cmidrule(lr){2-3} 
\cmidrule(lr){4-5} 
\cmidrule(lr){6-7} 
\cmidrule(lr){8-9} 
\cmidrule(lr){10-11}
& Default & Slerp 
& Default & Slerp 
& Default & Slerp 
& Default & Slerp 
& Default & Slerp \\
\midrule
Avg. 
& 31.46 & \textbf{89.12} 
& 34.29 & \textbf{90.38} 
& 50.04 & \textbf{89.54} 
& 50.65 & \textbf{90.25} 
& 54.83 & \textbf{88.60} \\
\midrule
T. 
& 31.88 & \textbf{99.67} 
& 32.87 & \textbf{99.67} 
& 76.83 & \textbf{98.69} 
& 76.17 & \textbf{99.67} 
& 58.92 & \textbf{99.18} \\
S. 
& 30.69 & \textbf{69.90} 
& 35.28 & \textbf{66.22} 
& 36.45 & \textbf{66.64} 
& 36.12 & \textbf{69.57} 
& 51.51 & \textbf{67.56} \\
V. 
& 32.66 & \textbf{88.36} 
& 35.17 & \textbf{99.51} 
& 41.79 & \textbf{95.72} 
& 40.42 & \textbf{94.91} 
& 50.36 & \textbf{89.81} \\
O. 
& 30.52 & \textbf{98.43} 
& 33.83 & \textbf{95.83} 
& 44.70 & \textbf{97.04} 
& 49.74 & \textbf{96.78} 
& 58.78 & \textbf{98.00} \\
\bottomrule
\end{tabular}
}
\end{table}

\begin{figure}[t]
    \centering
    \includegraphics[width=0.85\columnwidth]{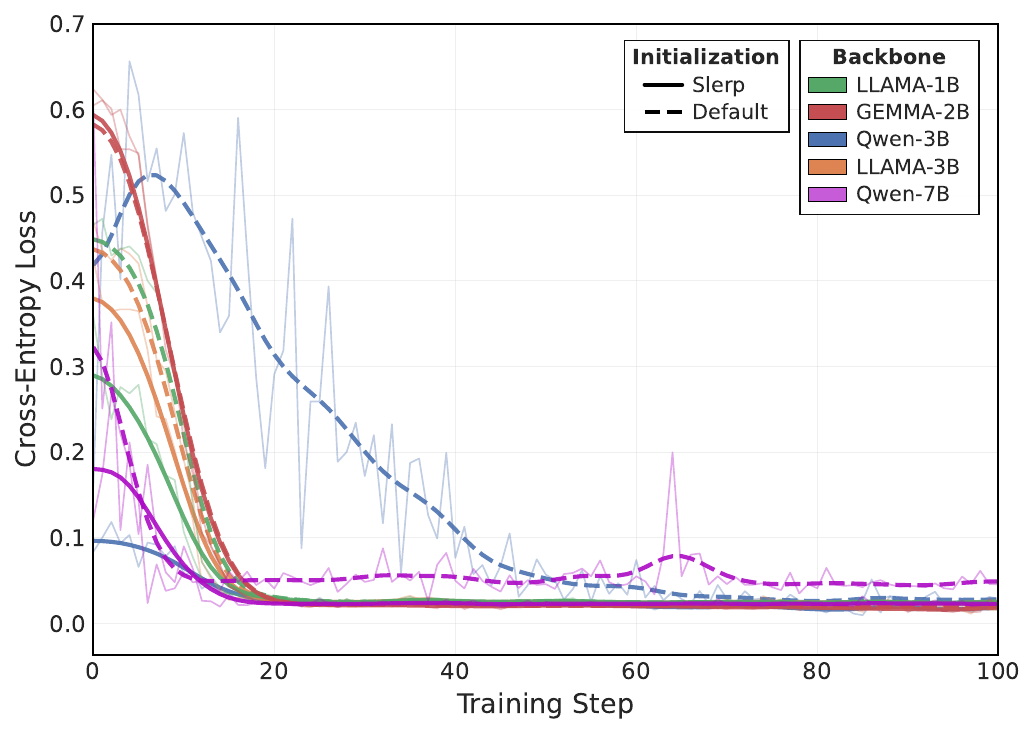}
    \caption{Training loss curves under Default and Slerp initialization schemes across different LLM backbones on TSQA dataset.}
    \label{fig:ce_loss_vs_steps_backbone}
\end{figure}

As presented in Table~\ref{tab:generalizability}, the \text{Slerp} method yields significant performance improvements over the \text{Default} baseline across all evaluated backbones on the TSQA dataset. Furthermore, the performance disparity between shape-dominant tasks and value-sensitive tasks remains consistent across different models, reinforcing the interpretability and efficacy of the COM strategy. In terms of training, the loss curves illustrated in Figure~\ref{fig:ce_loss_vs_steps_backbone} demonstrate that the ordered initialization consistently achieves a higher convergence rate compared to the disordered counterparts, regardless of the backbone (with Gemma as a minor exception).
This consistency aligns with the observations reported in Section~\ref{subsec:exp_hard_constraints}. Collectively, these results underscore the robust cross-architecture and cross-scale generalizability of COM, indicating that its benefits are generalizable across model families and parameter scales.

\vspace{0.5em}
\noindent\fbox{
\parbox{0.98\linewidth}{
\textbf{Finding 3:} The COM strategy exhibits robust generalizability, consistently achieving performance gains and maintaining interpretability across LLM backbones of different families and parameter scales.
}}

\subsection{RQ4: Universal Potential of COM}

\label{sec:universal_potential}

To investigate the potential of our COM-enhanced TS-LLMs as a universal model for diverse TSA tasks, we evaluate its performance across three fundamental tasks: question-answering, classification, and description. Following the best practices identified in previous analysis, we benchmark against established domain-specific baselines.

\paragraph{Question-answering (QA).}
We assess question-answering capabilities on the TSQA dataset. Table~\ref{tab:reasoning_results} presents the performance of various models. The empirical results demonstrate that our model significantly outperforms all baselines (TS-LLMs and general LLMs). Notably, at a 3B parameter scale, our model achieves a superior performance level, surpassing the state-of-the-art ChatTime by over 20\%. This underscores the critical role of the COM strategy in augmenting the question-answering capacities of TS-LLMs.

\begin{table}[htbp]
\centering
\caption{Accuracy (\%) comparison on the TSQA dataset, where Llama3 (8B), Qwen2.5 (3B), and DeepSeek-V4-Pro represent general LLMs with text-based TS inputs, while ChatTime and ChatTS represent TS-LLMs.}
\label{tab:reasoning_results}
\resizebox{\columnwidth}{!}{
\begin{tabular}{l|cccccc}
\toprule
 & \textbf{Ours} & \textbf{ChatTime} & \textbf{ChatTS} & \textbf{DeepSeek-V4} & \textbf{Llama3} & \textbf{Qwen2.5} \\ 
\midrule
T.    & \textbf{99.75} & \underline{87.89} & 82.91 & 79.59 & 28.57 & 35.66 \\
S.    & \textbf{95.82} & \underline{64.68} & 57.19 & 42.31 & 40.38 & 33.19 \\
V.    & \textbf{99.43} & 45.89 & 33.06 & \underline{46.30} & 38.89 & 37.75 \\
O.    & \textbf{99.48} & \underline{84.69} & 74.26 & 48.84 & 32.56 & 31.30 \\ 
\midrule
Avg.  & \textbf{98.62} & \underline{76.05} & 61.58 & 54.04 & 35.35 & 34.54 \\ 
\bottomrule
\end{tabular}
}
\end{table}

\paragraph{Classification.}
The classification capability of our model is evaluated on the RWC dataset (see Appendix~\ref{app:detail_dataset} for details). As shown in Table~\ref{tab:classification_results}, our COM-enhanced token-based TS-LLM achieves the highest accuracy and F1 score compared to task-specific classification baselines. These results indicate that the COM strategy helps effectively capture discriminative features.

\begin{table}[htbp]
\centering
\caption{Classification performance on the RWC dataset. Our model is compared against various task-specific baselines.}
\label{tab:classification_results}
\resizebox{\columnwidth}{!}{
\begin{tabular}{l|cc}
\toprule
\textbf{Model} & \textbf{Acc.} & \textbf{F1} \\ \midrule
FormerTime~\cite{cheng2023formertime}        & 0.7803 & 0.7796 \\
TCN~\cite{bai2018empirical-tcn}               & 0.7113 & 0.7109 \\
MiniROCKET~\cite{dempster2021minirocket}        & 0.7569 & 0.7556 \\
TimeMAE~\cite{cheng2023timemae}           & 0.7690 & 0.7664 \\
TS-TCC~\cite{eldele2021time-tstcc}            & 0.6979 & 0.6931 \\ 
InstructTime~\cite{cheng2025instructime}        & 0.7599 & 0.7578 \\
\midrule
\textbf{Ours (Slerp)} & \textbf{0.8135} & \textbf{0.8153} \\ \bottomrule
\end{tabular}
}
\end{table}


\paragraph{Description.}
We evaluate our model on four subsets of BEDTime, covering both synthetic and real-world scenarios with short and long textual descriptions (see Appendix~\ref{app:detail_dataset} for details). 
As shown in Table~\ref{tab:description_results}, our model consistently outperforms larger LLMs, VLMs, and TS-LLMs across a range of description-oriented metrics, including lexical overlap, sequence-level similarity, and semantic consistency. 
These results demonstrate that the COM strategy effectively enhances the model's ability to generate faithful time series descriptions and generalizes well across different data sources and description granularities.

\begin{table}[htbp]
\centering
\caption{Performance comparison on the BEDTime dataset. The best results are highlighted in bold, and the second-best results are underlined.}
\label{tab:description_results}
\resizebox{\columnwidth}{!}{
\begin{tabular}{l|cccccc}
\toprule
\textbf{Model} & \textbf{BLEU-2} & \textbf{BLEU-4} & \textbf{ROUGE-L} & \textbf{METEOR} & \textbf{DeBERTa F1} & \textbf{SimCSE Cos} \\
\midrule
Qwen2.5-32B           & 0.0623 & 0.0218 & 0.1624 & 0.1493 & 0.5255 & 0.5623 \\
Qwen3-32B             & 0.1156 & 0.0599 & 0.2316 & 0.2075 & \underline{0.5509} & \underline{0.6066} \\
Llama-3-11B-V         & 0.0871 & 0.0524 & 0.1920 & 0.1788 & 0.5269 & 0.5356 \\
GLM-5.1               & \underline{0.1222} & 0.0649 & 0.2291 & 0.1909 & 0.5388 & 0.5798 \\
DeepSeek-V4-Pro       & 0.1183 & \underline{0.0671} & \underline{0.2425} & \underline{0.2087} & 0.5506 & 0.5870 \\
ChatTS-14B            & 0.0832 & 0.0339 & 0.1794 & 0.1696 & 0.5173 & 0.5190 \\
\midrule
\textbf{COM (ours)}   & \textbf{0.2171} & \textbf{0.1357} & \textbf{0.3662} & \textbf{0.3082} & \textbf{0.6179} & \textbf{0.6388} \\
\bottomrule
\end{tabular}
}
\end{table}

Overall, these results across question-answering, classification, and description tasks suggest COM-enhanced TS-LLMs exhibit strong potential as a universal model for diverse TSA scenarios. Rather than being tailored to a single task type, the COM strategy consistently improves understanding, discrimination, and generation within the same token-based modeling paradigm.

\vspace{0.5em}
\noindent\fbox{
\parbox{0.95\linewidth}{
\textbf{Finding 4:} Our COM-enhanced TS-LLM exhibits significant universal potential, demonstrating competitive performance across TSA tasks compared to various baselines.
}}
\section{Conclusion}

In this paper, we address a critical limitation in current token-based TS-LLM paradigms: the neglect of intrinsic continuity and ordinality of time series tokens in the embedding space. To bridge this gap, we propose \textbf{COM} \textit{(Continuity and Ordinality Matter),} a continuity- and ordinality-aware strategy designed to enforce these geometric properties through a combination of hard and soft constraints. Experimental results demonstrate that COM substantially enhances time series modeling capabilities, accelerates model convergence, and exhibits robust generalizability across model architectures. Moreover, its competitive performance across diverse TSA tasks highlights its potential as a foundation for large-scale universal TS-LLMs. Beyond time series analysis, our findings may provide useful insights for modeling other forms of intrinsically ordered data.

\newpage

\section*{Limitations}

This work has several limitations. First, our study focuses on token-based TS-LLMs, where continuous time series values are quantized into discrete TS tokens and incorporated into the LLM vocabulary. Therefore, our conclusions may not directly transfer to other TS-LLM paradigms, such as alignment-based or vision-based approaches.

Second, although COM introduces geometric priors for TS token embeddings, we do not systematically characterize how different geometric properties affect downstream performance. Our experiments compare several initialization strategies with varying degrees of continuity and ordinality, leaving a more principled geometry--performance analysis for future work.

Third, the effectiveness of COM may depend on tokenization precision, normalization strategy, and regularization hyperparameters. In particular, finer quantization improves numerical resolution but increases vocabulary size, leading to a trade-off between modeling precision and computational cost.


Finally, while our experiments cover representative TSA tasks, the evaluation is still limited to selected benchmarks, model backbones, and data distributions. Real-world time series may involve stronger noise, missing values, irregular sampling, distribution shifts, or domain-specific constraints. Broader evaluations on larger models, longer sequences, and more diverse application domains are needed to further validate the generalizability of COM.

\section*{Ethical Considerations}

This work focuses on improving token-based TS-LLMs and is evaluated only on research benchmarks. It does not involve collecting personal data, human-subject annotations, or direct deployment in real-world decision-making systems. Therefore, we do not identify immediate ethical concerns in the current study. Nevertheless, since time series data may appear in sensitive domains such as healthcare, finance, and industrial monitoring, future deployment should ensure proper domain validation, privacy protection, and human oversight.





\bibliography{com}

\clearpage
\appendix
\section{Related Work}

\subsection{Time Series Large Language Models}

TS-LLMs leverage the question-answering and reasoning capabilities of LLMs for time series analysis (TSA) through four primary paradigms. \textbf{Text-based} methods (e.g., Time-MQA~\cite{kong2025time-time-mqa}, LLM-Time~\cite{gruver2023llmtime-zeroshot}) convert time series data into plain-text sequences, but often suffer from high amout of tokens and sensitivity to numerical precision. \textbf{Vision-based} methods (e.g., TAMA~\cite{zhuang2024see-tama}, Insight-Miner~\cite{zhang2023insight-miner}) plot time series as images to utilize vision-language models, though they are frequently constrained by image resolution and the high cost of visual tokens. \textbf{Alignment-based} methods (e.g., ChatTS~\cite{xie2025chatts}, ~\citep{chow2024towards-chow}) employ dedicated encoders to map temporal features into the LLM embedding space; however, their reliance on patch-based representations can obscure fine-grained local structures. \textbf{Token-based} methods (e.g., ChatTime~\cite{chattime,quinlan2026chat-ts,ansari2024chronos}) discretize values into an extended vocabulary, enabling efficient, unified modeling of text and time series. While promising, existing token-based methods typically treat TS tokens as normal vocabulary entries, overlooking their intrinsic continuity and ordinality. To address this, our COM strategy explicitly enforces these properties within the embedding space to better align with time series characteristics.

\subsection{Numerical and Continuous Embeddings}
Embedding techniques were originally introduced to map sparse, high-dimensional textual tokens into low-dimensional dense representations~\citep{word2vec}, and have therefore been particularly effective at modeling discrete concepts such as word pieces~\citep{bpe} and molecules composed of atoms~\citep{transformercpi}. However, in other domains (including time series analysis), many concepts exhibit partial orders or continuous measurements. Naively discretizing such concepts often destroys their inherent ordering and continuity. As a result, it is necessary to leverage continuous embedding techniques that preserve these structures. For example, in recommender systems, AutoDis~\citep{guo2021embedding} models continuous features (e.g., user age and item price) using continuous embeddings based on meta-embeddings and soft bucketization, ensuring that numerically similar values remain close in the embedding space. Similarly, prior work~\citep{gorishniy2022embeddings} extends continuous embedding methods to tabular data and demonstrates performance gains on tabular tasks such as click-through rate (CTR) prediction~\citep{pan2019warm}. Despite these advances, the application of continuous embeddings to time series data, which also exhibit partial ordering and continuous measurements, remains relatively underexplored in the context of TS-LLMs~\citep{davies2025language}.



\section{Detailed Introduction on TS-LLMs}
Time series large language models (TS-LLMs) are built upon large language models (LLMs) and leverage their natural language understanding and reasoning capabilities to address time series analysis tasks. Existing TS-LLMs can be broadly categorized into four paradigms based on how time series are integrated with LLMs: text-based, vision-based, alignment-based, and token-based approaches.

\textbf{Text-based methods} (e.g., Time-MQA~\cite{kong2025time-time-mqa}, LLM-Time~\cite{gruver2023llmtime-zeroshot}) convert time series into plain-text sequences and feed them into LLMs, with task-specific prompts used to solve different downstream tasks. While conceptually simple, these methods typically require substantial token overhead and are highly sensitive to the tokenizer design and numerical precision of different LLMs, resulting in limited numerical expressiveness and only preliminary performance.

\textbf{Vision-based methods} (e.g., TAMA~\cite{zhuang2024see-tama}, Insight-Miner~\cite{zhang2023insight-miner}) represent time series as images and rely on the visual understanding capabilities of vision-language models (VLMs) to perform analysis and reasoning. However, these methods are constrained by image resolution and the high cost of visual tokens, and its effectiveness heavily depends on the inherent ability of VLMs to capture temporal features and patterns.

\textbf{Alignment-based methods} (e.g., ChatTS\cite{xie2025chatts} and ~\citep{chow2024towards-chow}) introduce dedicated time series encoders and align the encoded time series features with the embedding space of LLMs. These methods often employ patch-based representations to reduce the number of input tokens, which limits the modeling of fine-grained local time series structures and poses challenges for recovering original numerical values required in precise reasoning tasks.

\textbf{Token-based methods} (e.g., ChatTime~\cite{chattime,quinlan2026chat-ts,ansari2024chronos}c) discretize continuous and unbounded time series values into a finite set of tokens and incorporate them as an extension of the LLM vocabulary, enabling unified sequence modeling of textual and TS tokens. With the rapid expansion of LLM context lengths (e.g., up to 128k tokens)~\cite{abouelenin2025phi-4-mini}, treating each numerical value as a token has become increasingly feasible, allowing this paradigm to strike a favorable balance between token efficiency and numerical quantization accuracy. Moreover, the natural interleaving of time series and text tokens offers the flexibility to support diverse time series analysis and reasoning tasks, making token-based approaches a promising direction toward general-purpose TS-LLMs.


\section{Implementation Details of Different initialization schemes}
\label{app:init_implement}

This appendix provides implementation details for the different initialization schemes used as hard constraints in our work, along with further conceptual explanations and illustrative pseudo-code examples.

\begin{figure*}[h]
  \centering

  \begin{subfigure}[t]{0.16\linewidth}
    \centering
    \includegraphics[width=\linewidth]{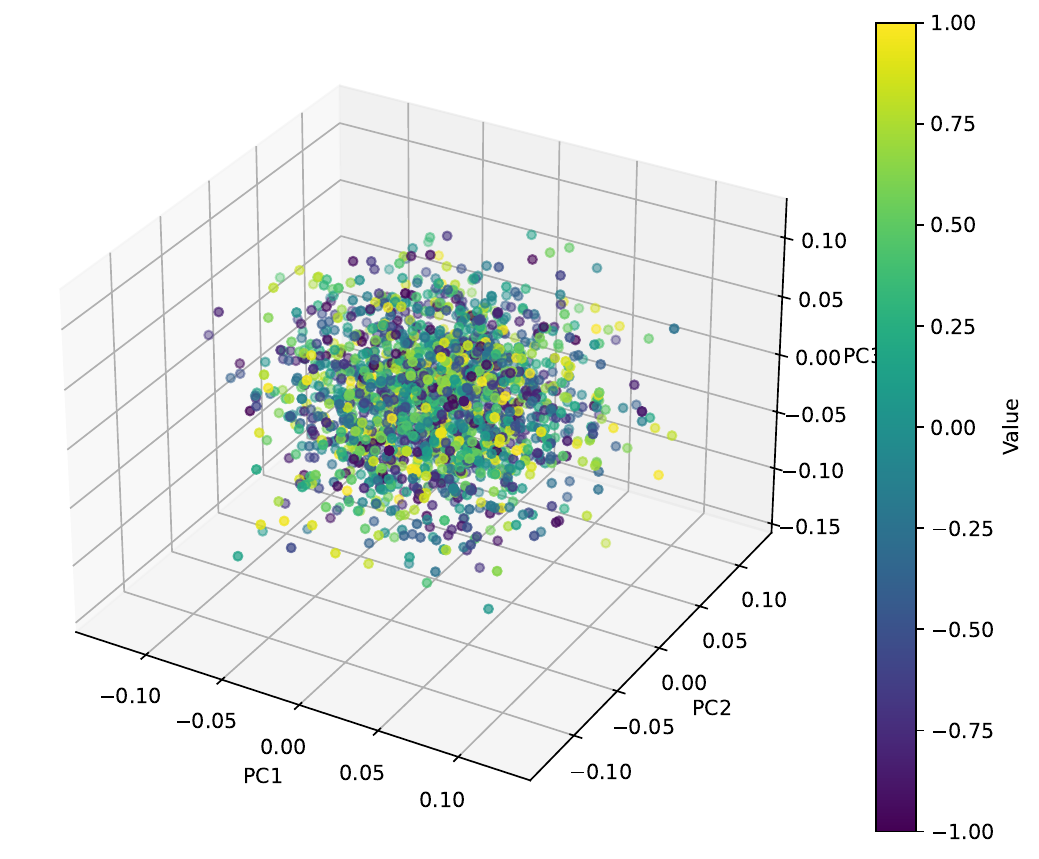}
    \caption{Default}
  \end{subfigure}
  \hfill
  \begin{subfigure}[t]{0.16\linewidth}
    \centering
    \includegraphics[width=\linewidth]{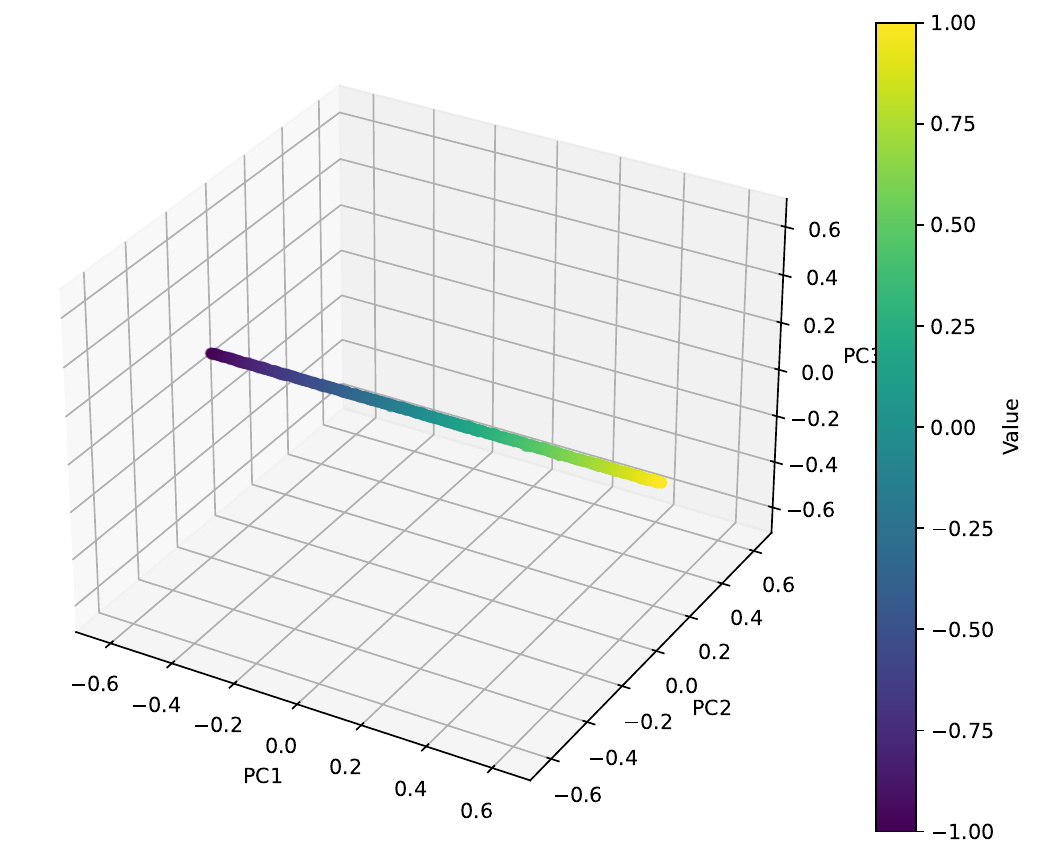}
    \caption{PCA-Main}
  \end{subfigure}
  \hfill
  \begin{subfigure}[t]{0.16\linewidth}
    \centering
    \includegraphics[width=\linewidth]{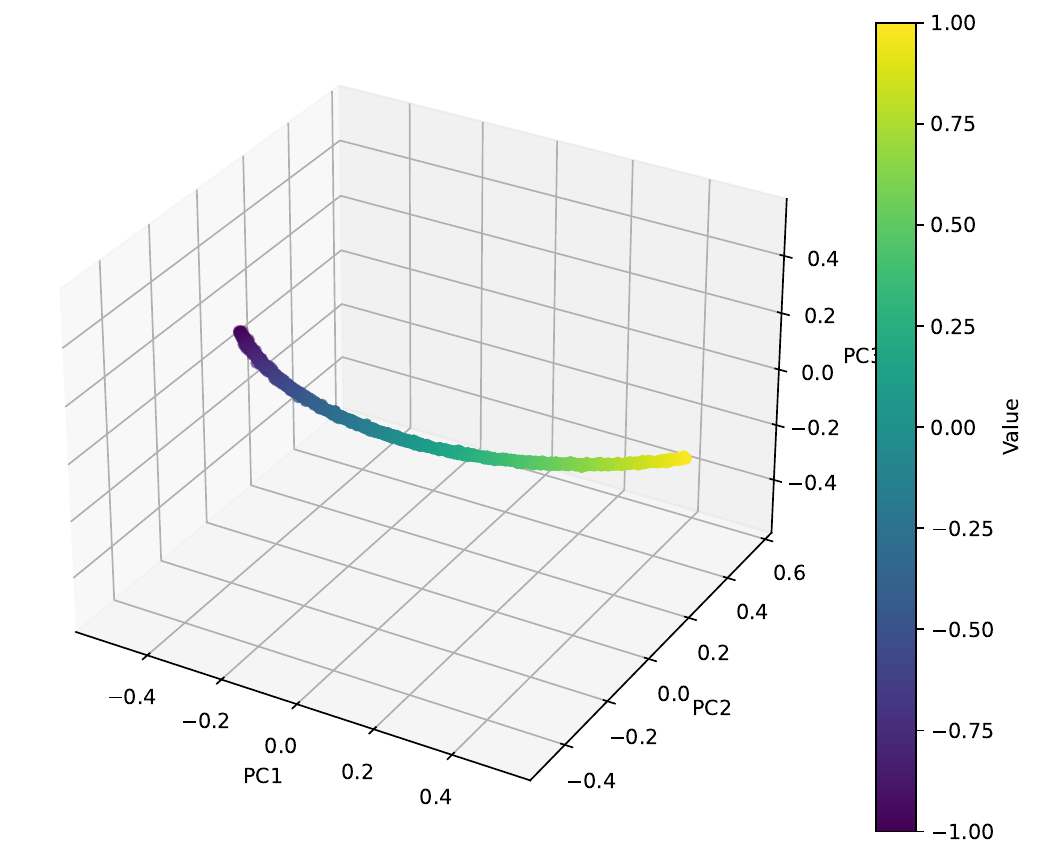}
    \caption{Slerp}
  \end{subfigure}
  \hfill
  \begin{subfigure}[t]{0.16\linewidth}
    \centering
    \includegraphics[width=\linewidth]{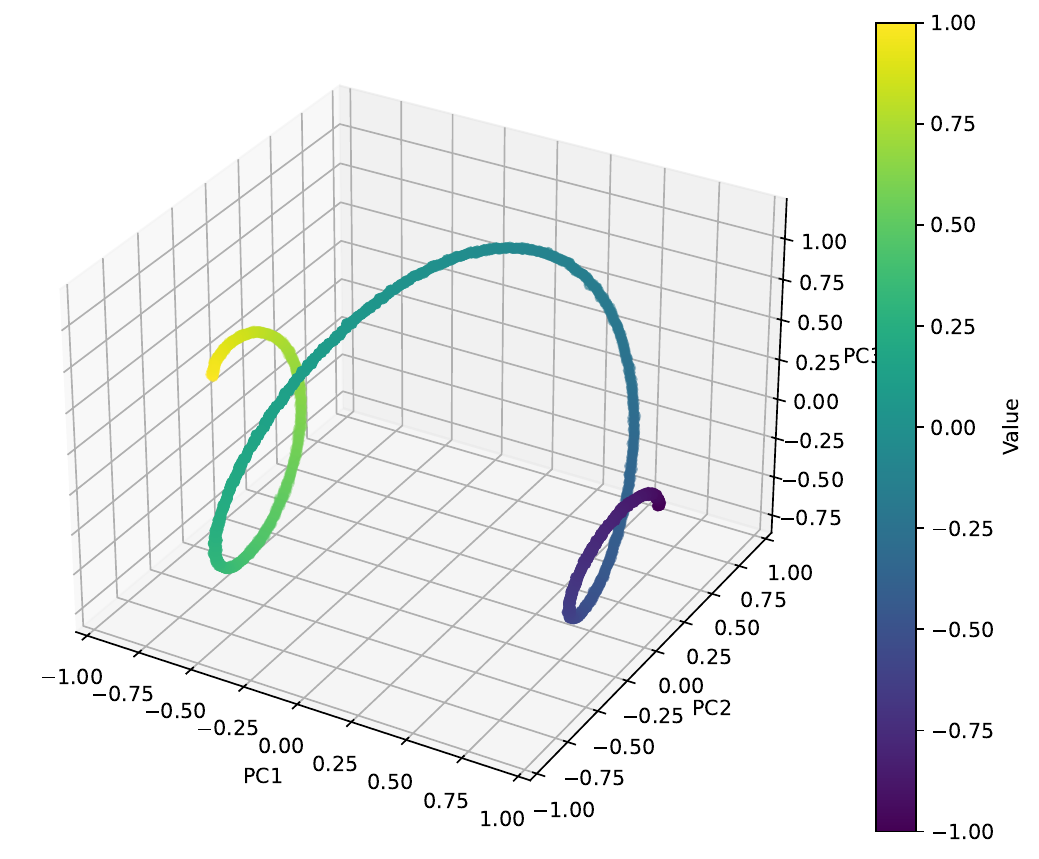}
    \caption{Helix-2}
  \end{subfigure}
  \hfill
  \begin{subfigure}[t]{0.16\linewidth}
    \centering
    \includegraphics[width=\linewidth]{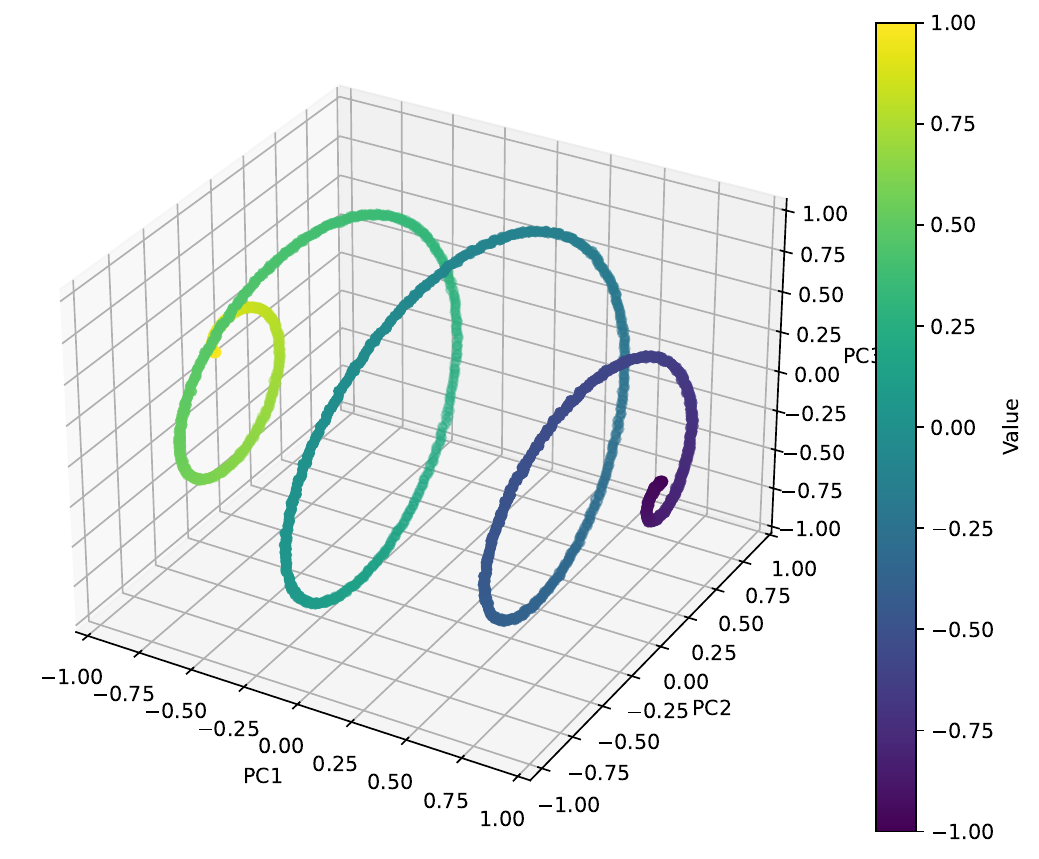}
    \caption{Helix-4}
  \end{subfigure}
  \hfill
  \begin{subfigure}[t]{0.16\linewidth}
    \centering
    \includegraphics[width=\linewidth]{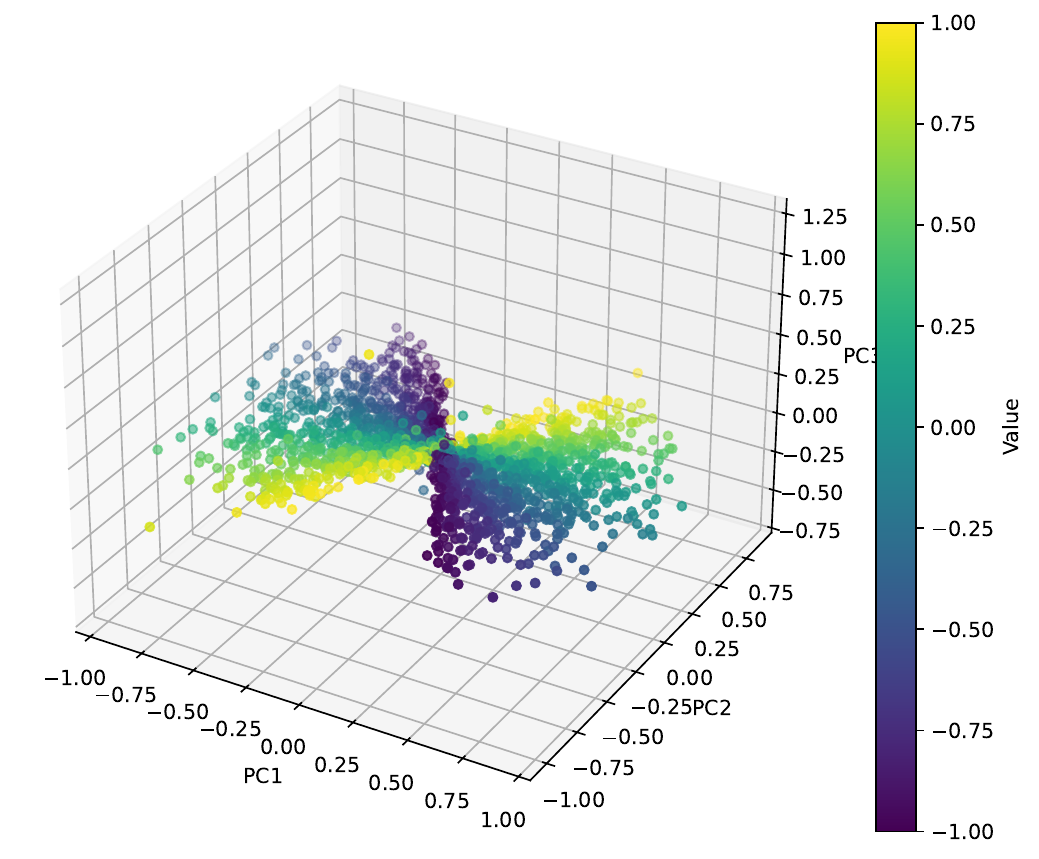}
    \caption{VMF}
  \end{subfigure}

  \caption{PCA-based 3D visualizations of TS embeddings under different embedding initialization schemes. The color from blue to yellow denotes time-series value tokens ranging from $-1$ to $+1$.
 (a) corresponds to the default Gaussian random initialization, while (b)–(f) represent our proposed prior-based initialization schemes.}
  \label{fig:visualization-3d}
\end{figure*}

\textbf{Default} strategy uses the Gaussian random initialization while expanding vocabulary size with the \texttt{model.resize\_token\_embeddings()} function.

\textbf{Slerp} strategy uses spherical linear interpolation to initialize TS token embeddings along a great-circle path between two randomly sampled points on the hypersphere. Small tangential and radial perturbations are added to increase diversity while preserving the global spherical geometry.

\begin{lstlisting}[
  language=Python,
  basicstyle=\ttfamily\small,
  caption={Pseudo-code for the Slerp initialization.},
  label={lst:slerp_great_circle},
  xleftmargin=0.02\linewidth,
  xrightmargin=0.02\linewidth,
  framexleftmargin=0.02\linewidth,
  framexrightmargin=0.02\linewidth
]
# Compute global embedding statistics
mean_embed  <- mean(embed_txt)
avg_radius  <- mean(||embed_txt - mean_embed||)

# Sample start and end points on the hypersphere
start_point <- random_unit_vector() * avg_radius
end_point   <- random_unit_vector() * avg_radius

for token index i from start_idx to end_idx:
    t <- (i - start_idx) / (num_tokens - 1)  # normalized position

    # Compute spherical linear interpolation (Slerp)
    omega <- arccos(dot(start_point, end_point) / avg_radius^2)
    point_on_sphere <- sin((1 - t) * omega) / sin(omega) * start_point +
                           sin(t * omega) / sin(omega) * end_point

    # Add tangential and radial perturbations
    radial_dir <- normalize(point_on_sphere)
    tangent_noise <- tangential_noise(radial_dir, scale=noise_scale * 
                    avg_radius)
    radial_noise  <- radial_noise_magnitude() * radial_dir
    final_point   <- point_on_sphere + tangent_noise/10 + radial_noise

    # Shift back to global mean
    embed[i] <- mean_embed + final_point
\end{lstlisting}

\textbf{PCA-Main} strategy uses the first principal component of the existing embeddings to place TS tokens along a line in the principal direction. Small tangential perturbations are added to increase diversity while preserving the global linear order.

\begin{lstlisting}[
  language=Python,
  basicstyle=\ttfamily\small,
  caption={Pseudo-code for the PCA-Main initialization.},
  label={lst:pca_main_init},
  xleftmargin=0.02\linewidth,
  xrightmargin=0.02\linewidth,
  framexleftmargin=0.02\linewidth,
  framexrightmargin=0.02\linewidth
]
# Compute global embedding statistics
mean_embed <- mean(embed_txt)
centered   <- embed_txt - mean_embed

# Compute principal component via PCA
principal_direction <- top_PCA_component(centered)

# Project embeddings onto the principal component
projections <- project(embed_txt, principal_direction)
(min_proj, max_proj) <- min_max(projections)
margin <- (max_proj - min_proj) * predefined_margin
new_min <- min_proj - margin
new_max <- max_proj + margin

for token index i from start_idx to end_idx:
    # Linear placement along principal component
    t <- new_min + (i - start_idx) / (num_tokens - 1) * (new_max - new_min)
    main_component <- t * principal_direction

    # Add small tangential noise
    noise <- tangential_noise(principal_direction, scale=noise_scale)

    # Shift back to global mean
    embed[i] <- mean_embed + main_component + noise
\end{lstlisting}

\textbf{Helix} strategy uses a spherical helical trajectory along the top three principal components of the existing embeddings. TS tokens are placed along the helix with latitude $\theta$ and longitude $\phi$, where $n$ of different variants Helix-$n$ specifies the number of helical turns. Small tangential and radial perturbations are applied to increase diversity while maintaining the overall helical structure.

\begin{lstlisting}[
  language=Python,
  basicstyle=\ttfamily\small,
  caption={Pseudo-code for the Helix initialization.},
  label={lst:helix_init},
  xleftmargin=0.02\linewidth,
  xrightmargin=0.02\linewidth,
  framexleftmargin=0.02\linewidth,
  framexrightmargin=0.02\linewidth
]
# Compute global embedding statistics
mean_embed <- mean(embed_txt)
avg_radius <- mean(||embed_txt - mean_embed||)

# Hyperparameters
num_turns  <- predefined number of helical turns
noise_scale <- small perturbation magnitude

# Obtain principal axes via PCA
(axis1, axis2, axis3) <- top 3 PCA components
axis2 <- orthogonalize(axis2, axis1)

for token index i from start_idx to end_idx:
    t <- (i - start_idx) / (num_tokens - 1)  # normalized position

    # Spherical angles for helical placement
    theta <- (t - 0.5) * pi
    phi   <- t * num_turns * 2 * pi

    # Compute embedding on the hypersphere
    embedding <- avg_radius * [
        cos(theta) * cos(phi) * axis1 +
        cos(theta) * sin(phi) * axis2 +
        sin(theta) * axis3 ]

    # Add tangential and radial perturbations
    radial_dir <- normalize(embedding)
    tangent_noise <- tangential_noise(radial_dir, scale=noise_scale * 
                    avg_radius)
    radial_noise  <- radial_noise_magnitude() * radial_dir
    final_embed   <- embedding + tangent_noise + radial_noise

    # Shift back to global mean
    embed[i] <- mean_embed + final_embed
\end{lstlisting}

\textbf{VMF} strategy places TS tokens along a spherical trajectory between two principal directions, sampling each token from a von Mises-Fisher (vMF)~\cite{mardia1975distribution-vmf} distribution around an interpolated mean direction. The sampled vectors are scaled to match the global radius distribution, preserving directional coherence.

\begin{lstlisting}[
  language=Python,
  basicstyle=\ttfamily\small,
  caption={Pseudo-code for the VMF initialization.},
  label={lst:vmf_init},
  xleftmargin=0.02\linewidth,
  xrightmargin=0.02\linewidth,
  framexleftmargin=0.02\linewidth,
  framexrightmargin=0.02\linewidth
]
# Compute global embedding statistics
mean_embed <- mean(embed_txt)
centered   <- embed_txt - mean_embed
radius_mean <- mean(||centered||)
radius_std  <- std(||centered||)

# Obtain start and end directions using PCA
start_direction <- top_PCA_component(centered)
end_direction   <- mix(top_PCA_components(centered, 2))
end_direction   <- normalize(end_direction)

for token index i from start_idx to end_idx:
    t <- (i - start_idx) / (num_tokens - 1)  # normalized position

    # Interpolate mean direction along spherical path
    omega <- arccos(dot(start_direction, end_direction))
    mu_direction <- sin((1-t)*omega)/sin(omega)*start_direction +
                        sin(t*omega)/sin(omega)*end_direction
    mu_direction <- normalize(mu_direction)

    # Sample from vMF distribution around mu_direction
    sample <- sample_vmf(mu_direction, concentration, embed_dim)

    # Adjust to appropriate radius
    target_radius <- sample_from_radius_distribution(radius_mean,radius_std)
    sample <- sample * target_radius

    # Shift back to global mean
    embed[i] <- mean_embed + sample
\end{lstlisting}

\section{Prompts and Templates}
\label{app:prompts_and_templates}

This appendix presents implementation details of prompts and templates in our work.
Figure~\ref{fig:template_example} illustrates how the TS-Processor converts time series data with heterogeneous lengths and numbers of channels into TS tokens and integrates them into the prompt template (as illustrated in Equation~\ref{eq:ts_processor}), including both the template format and an example. The following gray text boxes present the prompt templates used in our experiments on different datasets across tasks.

\begin{figure*}[h]
\centering
\begin{tcolorbox}[
    width=\textwidth,
    colback=gray!5,
    colframe=gray!70,
    title=TS-Processor Template and Example,
    fontupper=\scriptsize\ttfamily,
    fonttitle=\normalsize\bfseries,
    boxrule=0.5pt,
    arc=1mm
]

\textbf{[Template]} \\
{\scriptsize
Given {n} time series:\\
Time series \{i\} is of length \{len\}, with statistical information:\{max, min, mean, std, left, right, mid\}. \\
$\dots$ \\

We normalize all time series values into range [$-1.0$, $+1.0$], with the same scale factor \{factor\}. 
After scaling, the \{n\} time series are below ("\{sep\}" is the separator between values): \\
Time series \{i\} is: \{seq\_ts\_i\}. \\
$\dots$
}

\vspace{1.5em}

\textbf{[Example]} \\
{\scriptsize
Given 3 time series: \\
Time series 1 is of length 4, with statistical information:
\{max:0.038, min:-0.964, mean:-0.229, std:0.424, left:0.001, right:-0.964, mid:0.008\}. \\
Time series 2 is of length 8, with statistical information:
\{max:1.100, min:-0.964, mean:-0.092, std:0.613, left:0.001, right:-0.964, mid:1.100\}. \\
Time series 3 is of length 5, with statistical information:
\{max:0.038, min:-0.964, mean:-0.260, std:0.385, left:0.001, right:-0.383, mid:0.008\}. \\

We normalize all time series values into range [-1.0, +1.0], with the same scale factor 1.100. 
After scaling, the 3 time series are below ("|" is the separator between values): \\
Time series 1 is:
|$<$0.001$>$|$<$0.035$>$|$<$0.007$>$|$<$$-$0.876$>$|. \\
Time series 2 is:
|$<$0.001$>$|$<$0.035$>$|$<$0.007$>$|$<$$-$0.876$>$|$<$1.000$>$|$<$0.035$>$|$<$0.007$>$|$<$$-$0.876$>$|. \\
Time series 3 is:
|$<$0.001$>$|$<$0.035$>$|$<$0.007$>$|$<$$-$0.876$>$|$<$$-$0.348$>$|. \\
}

\end{tcolorbox}
\caption{Template and example of TS-Processor for time-series processing. 
The template consists of statistical metadata and quantized TS sequences. 
In the example, time series values are normalized to $[-1, +1]$ and enclosed by ``\textless'' and ``\textgreater'', with ``\texttt{|}'' used as the separator between TS tokens. 
The example demonstrates that the proposed method inherently supports multivariate inputs with heterogeneous sequence lengths. 
The resulting $S_T$ serves as the time series information in the full prompt for TS-LLMs.
}
\label{fig:template_example}
\end{figure*}

\begin{figure*}[h]
\centering
\begin{tcolorbox}[
    width=\textwidth,
    colback=gray!5,
    colframe=gray!70,
    title=Prompt Template Used on TSQA Dataset,
    fontupper=\small\ttfamily,
    fonttitle=\normalsize\bfseries,
    boxrule=0.5pt,
    arc=1mm
]
\#\# Time Series Information:

<Insert the time series context here, e.g., statistical metadata, quantized sequences, etc.>\\

\#\# Question:

<Insert the question here>\\

\#\# Options:

<Insert all candidate options here, exactly as listed>\\

\#\# Instruction:

Output the option letter followed by the full option text exactly as listed in Options. Do not add explanations or generate any additional content.
\end{tcolorbox}
\end{figure*}

\begin{figure*}[h]
\centering
\begin{tcolorbox}[
    width=\textwidth,
    colback=gray!5,
    colframe=gray!70,
    title=Prompt Template Used on MMTS-InWild Subset,
    fontupper=\small\ttfamily,
    fonttitle=\normalsize\bfseries,
    boxrule=0.5pt,
    arc=1mm
]
\#\# Time Series Information:

<Insert the time series context here, e.g., statistical metadata, quantized sequences, etc.>\\

<Insert any auxiliary information here, if applicable>\\

\#\# Question:

<Insert the question here>\\

\#\# Options:

<Insert all candidate options here, exactly as listed>\\

\#\# Instruction:

Output the option letter followed by the full option text exactly as listed in Options. Do not add explanations or generate any additional content.
\end{tcolorbox}
\end{figure*}

\begin{figure*}[h]
\centering
\begin{tcolorbox}[
    width=\textwidth,
    colback=gray!5,
    colframe=gray!70,
    title=Prompt Template Used on TimeSeriesExam Dataset,
    fontupper=\small\ttfamily,
    fonttitle=\normalsize\bfseries,
    boxrule=0.5pt,
    arc=1mm
]
\#\# Time Series Information:

<Insert the time series context here, e.g., statistical metadata, quantized sequences, etc.>\\

\#\# Question:

<Insert the question here>\\

\#\# Options:

<Insert all candidate options here, exactly as listed>\\

\#\# Instruction:

For a given question, you should exactly choose one answer from the options, and output the full answer. Don't generate anything else.
\end{tcolorbox}
\end{figure*}

\begin{figure*}[h]
\centering
\begin{tcolorbox}[
    width=\textwidth,
    colback=gray!5,
    colframe=gray!70,
    title=Prompt Template Used on BEDTime Dataset,
    fontupper=\small\ttfamily,
    fonttitle=\normalsize\bfseries,
    boxrule=0.5pt,
    arc=1mm
]
\#\# Time Series Information:

<Insert the time series context here, e.g., statistical metadata, quantized sequences, etc.>\\

\#\# Question:

<Insert the question here>\\

\#\# Instruction:

Output a brief natural-language description of the given time series. Do not generate analysis or any additional content.
\end{tcolorbox}
\end{figure*}

\begin{figure*}[h]
\centering
\begin{tcolorbox}[
    width=\textwidth,
    colback=gray!5,
    colframe=gray!70,
    title=Prompt Template Used on Forecasting Tasks,
    fontupper=\small\ttfamily,
    fonttitle=\normalsize\bfseries,
    boxrule=0.5pt,
    arc=1mm
]
\#\# Time Series Information:\par
\smallskip
\textless Insert the historical time series context here, e.g., statistical metadata and quantized historical sequence.\textgreater

\bigskip

\#\# Context Information:\par
\smallskip
\textless Insert additional context information here, e.g., dataset-specific background or characteristics.\textgreater

\bigskip

\#\# Instruction:\par
\smallskip
Given a univariate historical time series of length \textless hist\_len\textgreater, your task is to predict the next \textless pred\_len\textgreater{} consecutive values.\par
Additional context information is provided to help you better understand the characteristics or background of the time series. You may use this context as auxiliary information, but your prediction must remain consistent with the historical data.

\smallskip

Requirements:\par
- Predict exactly \textless pred\_len\textgreater{} numerical values.\par
- The prediction should be based only on the provided historical time series.\par
- Do NOT include explanations, analysis, or any additional text.\par
- Output must strictly follow the format below.

\smallskip

Output format:\par
The predicted next \textless pred\_len\textgreater{} time series values are (scale \textless scale\textgreater):\par
[Your Answer]
\end{tcolorbox}
\end{figure*}

\section{Experimental Setup}
\label{app:exp_setup}

\textbf{Training Configuration.}
As described in Sec.~\ref{subsec:training}, our training procedure follows a two-phase pipeline: \textit{self-distillation alignment} followed by \textit{downstream fine-tuning}. For mechanistic and ablation studies of the \textit{COM} strategy (RQ1--RQ3), we primarily use only the second-phase fine-tuning to isolate the effect of the investigated components. The full two-phase pipeline is used for evaluating peak performance and generalizability (RQ4).

By default, we adopt \texttt{Qwen2.5-3B-Instruct} as the LLM backbone~\cite{qwen2025qwen25technicalreport-qwen2.5}. The precision $\epsilon$ of the TS-Processor is set to $0.001$, yielding a TS-token vocabulary of $|\mathcal V_{\mathrm{ts}}| = 2001$. This vocabulary size is chosen to be compatible with the original vocabulary size of the backbone, i.e., $|\mathcal V_{\mathrm{llm}}| = 151{,}936$. For token formatting, we use the wrapper \texttt{"<{}<{}<\{num\}>{}>{}"} and adopt \texttt{"|"} as the separator, which improves readability and preserves clear token boundaries.

We perform full-parameter fine-tuning with the AdamW optimizer using a learning rate of $1\times 10^{-5}$, a cosine learning-rate scheduler with a warmup ratio of $0.03$, and a weight decay of $0.01$. All experiments are conducted on a single NVIDIA A800 80GB GPU with a total batch size of $64$, implemented as a per-device batch size of $16$ with $4$ gradient accumulation steps, for one training epoch.

Unless otherwise specified, all main results are reported from a single
run with random seed 42. We choose this reporting protocol due to the
large computational cost of full-parameter fine-tuning on TS-LLMs across multiple datasets and task settings. To ensure
reproducibility, we fix all random seeds, training configurations, data
splits, and evaluation protocols throughout the experiments.

To assess potential sensitivity to random initialization and data-ordering
effects, we additionally conduct experiments with alternative random seeds
on representative settings. The observed trends are generally consistent
across seeds. Unless otherwise specified, we report the seed-42 results in
the main tables.

\textbf{Datasets and Tasks.}
We evaluate our model across a broad spectrum of time series analysis (TSA) tasks. For fundamental time series question answering, we use TSQA~\cite{chattime}, which covers four basic properties, including Trend, Seasonality, Volatility, and Outliers, over univariate sequences of lengths $\{64,128,256,512\}$. We train on the ChatTime fine-tuning data~\cite{chattime} for TSQA-related experiments. For complex time series reasoning, we adopt the MMTS-InWild subset from MMTS-Bench~\cite{anonymous2026mmts-bench}, which evaluates diverse analysis and reasoning abilities under a hierarchical taxonomy; following the MMTS synthesis protocol, we generate approximately 2.5k fine-tuning samples on heterogeneous time-series data. To further assess general-purpose TSA capabilities, we include the RWC~\cite{whale_detection_challenge-rwc} dataset for classification, BEDTime~\cite{sen2025bedtime} for time-series description generation, and CGTSF-PTF, ETT (h1, h2, m1, m2), Traffic, and Weather for forecasting. Detailed dataset information is provided in Appendix~\ref{app:detail_dataset}, and task modeling and prompt designs are provided in Appendix~\ref{app:prompts_and_templates}.


\section{Implementation of Soft Constraints}
Algorithm~\ref{alg:com} presents the pseudo-code details for implementing our regularization losses, which serve as soft constraints in COM.

\begin{algorithm}[h!]
\caption{Implementation of regularization losses (soft constraints).}
\label{alg:com}
\small
\begin{algorithmic}[1]
\STATE \textbf{Input:} TS embeddings $\mathbf{E}_{ts} \in \mathbb{R}^{N \times D}$, step size $k$, margins $m_{\text{ord}}, m_{\text{mono}}$
\STATE \textbf{Output:} Ordinality loss $\mathcal{L}_{\text{ord}}$, monotonicity loss $\mathcal{L}_{\text{mono}}$

\STATE $\mathbf{P} \leftarrow \text{PCA}(\mathbf{E}_{ts}, \text{dim}=3)$
\STATE $\mathbf{E}' \leftarrow \mathbf{E}_{ts}\mathbf{P}$ \COMMENT{$\mathbf{E}' \in \mathbb{R}^{N \times 3}$}

\STATE \textbf{Compute Ordinality Loss:}
\STATE $\mathbf{E}_{i} \leftarrow \mathbf{E}'[2k:]$ \COMMENT{$\mathbf{e}'_i$}
\STATE $\mathbf{E}_{i-k} \leftarrow \mathbf{E}'[k:-k]$ \COMMENT{$\mathbf{e}'_{i-k}$}
\STATE $\mathbf{E}_{i-2k} \leftarrow \mathbf{E}'[:-2k]$ \COMMENT{$\mathbf{e}'_{i-2k}$}
\STATE $d_{\text{near}} \leftarrow \|\mathbf{E}_{i} - \mathbf{E}_{i-k}\|_2$
\STATE $d_{\text{far}} \leftarrow \|\mathbf{E}_{i} - \mathbf{E}_{i-2k}\|_2$
\STATE $\mathcal{L}_{\text{ord}} \leftarrow \text{mean}\left(\text{ReLU}(d_{\text{near}} - d_{\text{far}} - m_{\text{ord}})\right)$

\STATE \textbf{Compute Monotonicity Loss:}
\STATE $\mathbf{E}_{i-k} \leftarrow \mathbf{E}'[:-2k]$ \COMMENT{$\mathbf{e}'_{i-k}$}
\STATE $\mathbf{E}_{i} \leftarrow \mathbf{E}'[k:-k]$ \COMMENT{$\mathbf{e}'_i$}
\STATE $\mathbf{E}_{i+k} \leftarrow \mathbf{E}'[2k:]$ \COMMENT{$\mathbf{e}'_{i+k}$}
\STATE $\Delta_{\text{prev}} \leftarrow \mathbf{E}_{i} - \mathbf{E}_{i-k}$
\STATE $\Delta_{\text{next}} \leftarrow \mathbf{E}_{i+k} - \mathbf{E}_{i}$
\STATE $sim \leftarrow \text{CosineSimilarity}(\Delta_{\text{prev}}, \Delta_{\text{next}})$
\STATE $\mathcal{L}_{\text{mono}} \leftarrow \text{mean}\left(\text{ReLU}(-sim - m_{\text{mono}})\right)$

\STATE \textbf{Compute Evaluation Metrics:}
\STATE $\mathcal{R}_{\text{ord}} \leftarrow \text{mean}\left(\text{ReLU}(d_{\text{near}} - d_{\text{far}})\right)$
\STATE $\mathcal{R}_{\text{mono}} \leftarrow \text{mean}\left(\text{ReLU}(-sim)\right)$

\STATE \textbf{Return:} $\mathcal{L}_{\text{ord}}, \mathcal{L}_{\text{mono}}, \mathcal{R}_{\text{ord}}, \mathcal{R}_{\text{mono}}$
\end{algorithmic}
\end{algorithm}

\section{Detailed Information on Datasets}
\label{app:detail_dataset}

This appendix documents the dataset used in our experiments, including their task formats, data modalities, application domains, and language coverage. Overall, the benchmarks cover diverse time series analysis scenarios, including question answering, reasoning, classification, forecasting, description generation, and multimodal reasoning, with textual components primarily in English.

\textbf{TSQA.} The TSQA (Time Series Question Answering)~\cite{chattime} dataset is a multimodal multi-choice question dataset introduced to assess the translation of time series data into natural language. Containing four distinct subtasks (Trend, Seasonality, Volatility, and Outliers), TSQA specifically targets the TS-to-text task by requiring models to interpret and identify specific patterns within time series data, such as distinguishing between constant, upward, and downward trends. The dataset presents time series segments alongside multiple-choice textual descriptions, challenging the model to comprehend the time series and select the correct linguistic explanation, thereby serving as a robust benchmark for validating the modality alignment and reasoning proficiency of unified time series foundation models like ChatTime. In our experiments, TSQA contains 48,000 samples in total, split into 38,400 training samples, 4,800 validation samples, and 4,800 test samples, following an 8:1:1 ratio.

\textbf{MMTS-Bench.} MMTS-Bench~\cite{anonymous2026mmts-bench} is a comprehensive multimodal benchmark designed to systematically evaluate the capabilities of Large Language Models (LLMs) in time series understanding and reasoning. Built upon a hierarchical taxonomy that spans feature analysis, temporal reasoning, and cross-modal alignment, the dataset comprises a total of 2,424 Time Series Question Answering (TSQA) pairs distributed across four distinct subsets: Base, InWild, Match, and Align. Specifically, the InWild subset serves as a critical component for assessing model performance in realistic, open-ended scenarios, consisting of 1,084 samples sourced from diverse real-world domains including finance, healthcare, climate, and cloud operations. This subset is constructed using a progressive real-world QA framework to ensure high data quality and relevance, providing a rigorous testbed for measuring the robustness and generalization ability of models when handling complex, authentic time series data.

\textbf{RWC.} The Right Whale Calls (RWC)~\cite{whale_detection_challenge-rwc} dataset is derived from the Marinexplore and Cornell University Whale Detection Challenge on Kaggle and is designed for automated detection of North Atlantic right whale ( Eubalaena glacialis ) vocalizations from underwater acoustic recordings. It consists of 2-second audio clips sampled at 2 kHz, collected by autonomous hydrophone buoys in the North Atlantic. The task is formulated as a binary classification problem that identifies the presence of right whale “up-calls,” a characteristic low-frequency (60–250 Hz), approximately one-second rising tonal call. The dataset includes 30,000 labeled training samples and 54,503 unlabeled test samples, and is challenging due to strong interference from ambient ocean noise and anthropogenic acoustic sources.

\textbf{CGTSF-PTF.} The CGTSF-PTF dataset~\cite{chattime} serves as a benchmark for context-guided time series forecasting, featuring 12 months of traffic flow data collected from 32 traffic detectors in Paris throughout 2012. Sampled at a one-hour granularity, this dataset aligns numerical time series records with auxiliary textual information—including dataset background, Open-Meteo~\cite{openmeteo2021} weather forecasts (such as temperature and weather codes), and calendar details like the day of the week and holidays—concatenated into coherent descriptions to enable multimodal modeling without future data leakage. Specifically, the forecasting task involves using historical sequence lengths of $\{48, 72, 96, 120\}$ to predict the traffic flow for the subsequent 24 time steps. In our experiments, CGTSF-PTF contains 46,080 samples in total, split into 36,864 training samples, 4,608 validation samples, and 4,608 test samples, following an 8:1:1 ratio.

\textbf{BEDTime.} BEDTime~\cite{sen2025bedtime} is a unified benchmark for evaluating models' ability to describe the structural properties of time series in natural language. It reformulates multiple public time series description datasets into a consistent evaluation framework covering description recognition, description differentiation, and open-ended description generation. In this work, we mainly use four representative subsets from BEDTime: SUSHI, TaxoSynth, TRUCE-Synthetic, and TRUCE-Stock, containing 1,400, 1,400, 1,677, and 5,687 samples, respectively. Each subset follows an approximately 9:1 train/test split. These subsets provide broad coverage across both synthetic and real-world scenarios, as well as short and long textual descriptions. Specifically, TRUCE-Stock contains short, noisy real-world stock price sequences paired with crowd-sourced descriptions, while TRUCE-Synthetic provides short synthetic sequences with human or template-based descriptions. In contrast, SUSHI contains long synthetic time series with prominent trends, seasonality, and noise, and TaxoSynth includes variable-length synthetic time series annotated with taxonomic descriptions of common time series patterns. Together, these subsets offer a diverse testbed for evaluating whether models can align time series patterns with natural language descriptions across different data sources, time series lengths, and linguistic complexities.

\section{Evaluation Metrics}
\label{sec:appendix-metrics}

This appendix provides the concrete instantiations of the task-specific
evaluation metric $\mathcal{M}$ introduced in Sec.~\ref{sec:preliminary}.
As defined in the preliminary formulation, each sample in a dataset
$\mathcal{D}$ consists of a time series input $T_i$, a question $Q_i$, and the corresponding ground-truth answer $G_i$. Given
the model prediction $A_i$, the overall performance is computed by
averaging $\mathcal{M}(G_i, A_i)$ over all $N$ test samples, as shown
in Equation~\ref{eq:perf}. Since $\mathcal{M}$ depends on the tasks,
we specify below the exact metrics used for multiple-choice question-answering and resaoning,
classification, forecasting, and description tasks in our
experiments.

\subsection{Question Answering and Reasoning}
\label{ssec:appendix-metrics-qa}

For multiple-choice QA and reasoning, $\mathcal{M}$
reduces to label agreement and we report \emph{accuracy},
\begin{equation}
\mathrm{Acc}(\mathcal{D})
\;=\; \frac{1}{N}\sum_{i=1}^{N}\mathbbm{1} \!\left[A_i = G_i\right].
\end{equation}

\subsection{Classification}
\label{ssec:appendix-metrics-cls}

For time series classification with $C$ classes, we evaluate model
predictions using accuracy and macro-averaged $F_1$ score. Accuracy
measures the overall proportion of correctly predicted samples, while
Macro-$F_1$ assigns equal weight to each class and is therefore more
robust to class imbalance. For class $c$, let $\mathrm{TP}_c$ denote
the number of samples correctly predicted as class $c$, $\mathrm{FP}_c$
the number of samples incorrectly predicted as class $c$, and
$\mathrm{FN}_c$ the number of samples from class $c$ incorrectly
predicted as other classes. The class-wise precision and recall are
then defined as
\begin{equation}
\begin{gathered}
P_c = \frac{\mathrm{TP}_c}{\mathrm{TP}_c + \mathrm{FP}_c},
\quad
R_c = \frac{\mathrm{TP}_c}{\mathrm{TP}_c + \mathrm{FN}_c}, \\[3pt]
\mathrm{Macro}\text{-}F_1
= \frac{1}{C}\sum_{c=1}^{C}\frac{2\,P_c\,R_c}{P_c + R_c}.
\end{gathered}
\end{equation}

\subsection{Time Series Description}
\label{ssec:appendix-metrics-cap}

For free-form time series description, we adopt a panel of complementary
metrics to evaluate generation quality from multiple perspectives. We
report metrics covering three aspects: \emph{lexical overlap}
(BLEU-$n$), \emph{sequence-level surface similarity}
(ROUGE-L and METEOR), and \emph{semantic consistency} (DeBERTa $F_1$
and SimCSE cosine similarity). Let $G = (g_1,\dots,g_{|G|})$ denote the
reference description and $A = (a_1,\dots,a_{|A|})$ denote the generated
answer. All metrics are computed for each sample and then uniformly
averaged over the test set $\mathcal{D}$.

\paragraph{BLEU-$n$.}
We report sentence-level BLEU-2 and BLEU-4~\cite{papineni-etal-2002-bleu} with uniform weights
$w_k = 1/n$, and apply Chen and Cherry's smoothing method~1
\citep{chen-cherry-2014-systematic} to avoid zero precision for
higher-order $n$-grams. Let $\mathcal{N}_k(\cdot)$ denote the
multiset of $k$-grams and $c_\bullet(\omega)$ denote the count of an
$n$-gram $\omega$. The BLEU-$n$ score is defined as
\begin{equation}
\begin{gathered}
\mathrm{BLEU}\text{-}n
\;=\; \mathrm{BP}\cdot
\exp\!\Bigl(\frac{1}{n}\sum_{k=1}^{n}\log p_k\Bigr), \\[3pt]
p_k
= \frac{
\sum_{\omega\in\mathcal{N}_k(A)}
\min\!\bigl(c_A(\omega),c_G(\omega)\bigr)
}{
\sum_{\omega\in\mathcal{N}_k(A)} c_A(\omega)
}.
\end{gathered}
\end{equation}
with brevity penalty
$\mathrm{BP} = \min\!\bigl(1, e^{1-|G|/|A|}\bigr)$.

\paragraph{ROUGE-L.}
ROUGE-L evaluates sequence-level overlap using the longest common
subsequence between the reference $G$ and the generated answer $A$
\citep{lin-2004-rouge}. We report its $F$-measure form with
$\beta = 1$:
\begin{equation}
\mathrm{ROUGE}\text{-}L
\;=\; \frac{2\,\mathrm{LCS}(G,A)}{|G| + |A|}.
\end{equation}
where $\mathrm{LCS}(G,A)$ denotes the length of the longest common
subsequence between the reference $G$ and the generated answer $A$.
We compute ROUGE-L using the \texttt{rouge\_score} package with
Porter stemming~\cite{porter1980algorithm}, which reduces the effect of simple morphological
variation, such as \emph{rises} and \emph{rising}.

\paragraph{METEOR.}
METEOR aligns hypothesis tokens to reference tokens through a cascade
of exact, stem, and synonym matches~\cite{lavie-agarwal-2007-meteor}.
Let $\mathfrak{A}$ be the resulting alignment, $\mathrm{ch}$ denote
the number of contiguous matched chunks, and define
$P = |\mathfrak{A}|/|A|$ and $R = |\mathfrak{A}|/|G|$. The METEOR
score is computed as
\begin{equation}
\resizebox{\columnwidth}{!}{$
\mathrm{METEOR}
\;=\; \Bigl(1 - \gamma\bigl(\mathrm{ch}/|\mathfrak{A}|\bigr)^{\theta}\Bigr)
       \cdot \frac{P\,R}{\alpha P + (1-\alpha) R}
$}
\end{equation}
with NLTK defaults $(\alpha,\gamma,\theta)=(0.9,\,0.5,\,3)$. The
fragmentation penalty favors more contiguous alignments, while the
stem- and synonym-aware matching makes METEOR less sensitive to
surface-form variation than exact $n$-gram matching.

\paragraph{DeBERTa $F_1$.}
We instantiate BERTScore with
\texttt{deberta-large-mnli} as the backbone~\cite{zhang-etal-2020-bertscore}, taking contextual token
embeddings from layer~9 following the library default. This backbone
is based on DeBERTa~\cite{he-etal-2021-deberta} and is fine-tuned on
MultiNLI~\cite{williams-etal-2018-broad}. For every reference token
$g_i$ and answer token $a_j$, let
$s_{ij} = \cos\!\bigl(\mathbf{e}(g_i),\,\mathbf{e}(a_j)\bigr)$ denote
their cosine similarity. The BERTScore $F_1$ is obtained by greedy
maximum matching:
\begin{equation}
\resizebox{\columnwidth}{!}{$
\begin{gathered}
F_{\mathrm{BS}}
\;=\; \frac{2\, P_{\mathrm{BS}}\, R_{\mathrm{BS}}}
            {P_{\mathrm{BS}} + R_{\mathrm{BS}}}, \\[2pt]
P_{\mathrm{BS}} = \tfrac{1}{|A|}\textstyle\sum_{j}\max_{i} s_{ij},
\;\;
R_{\mathrm{BS}} = \tfrac{1}{|G|}\textstyle\sum_{i}\max_{j} s_{ij}.
\end{gathered}
$}
\end{equation}
We report $F_{\mathrm{BS}}$ without IDF weighting or baseline rescaling.
That is, all tokens contribute equally, and no model-specific baseline
calibration is applied. Since these configuration choices affect the
absolute scale of BERTScore, we use this metric only for relative
comparison among systems evaluated under the same setting.

\paragraph{SimCSE cosine.}
For sentence-level semantic similarity, we encode each text using
\texttt{sup-simcse-roberta-large}~\cite{gao-etal-2021-simcse}. We
take the \texttt{pooler\_output} and apply $\ell_2$ normalization; let
$\hat{\mathbf{e}}(\cdot)$ denote the resulting unit vector. The SimCSE
score is the cosine similarity between reference and hypothesis
embeddings:
\begin{equation}
\mathrm{SimCSE}(G, A)
\;=\; \hat{\mathbf{e}}(G)^{\!\top}\,\hat{\mathbf{e}}(A),
\end{equation}
which is then averaged over $\mathcal{D}$. SimCSE complements
token-level matching metrics by providing a sentence-level semantic
similarity measure.

\paragraph{Implementation details.}
All automatic evaluation metrics are computed at the sample level and
then uniformly averaged over the test set. For lexical and surface-form
metrics, both references and generated answers are lowercased and
tokenized by whitespace splitting. We do not remove punctuation. To
avoid degenerate overly long generations causing memory issues for
embedding-based metrics, both references and generated answers are
truncated to at most 200 whitespace-separated words before evaluation.

We compute BLEU scores using
\texttt{nltk.translate.bleu\_score.sentence\_bleu} with
\texttt{SmoothingFunction().method1}. BLEU-2 and BLEU-4 use uniform
$n$-gram weights, i.e., $(1/2,1/2)$ and $(1/4,1/4,1/4,1/4)$,
respectively~\cite{bird2006nltk}. ROUGE-L is computed using the
\texttt{rouge\_score} package with
\texttt{rouge\_scorer.RougeScorer(["rougeL"], use\_stemmer=True)},
and we report the $F$-measure. METEOR is computed using
\texttt{nltk.translate.meteor\_score.meteor\_score} with the NLTK
default parameters.

For DeBERTa $F_1$, we instantiate BERTScore using the
\texttt{bert\_score} package with \texttt{deberta-large-mnli} as the
backbone, \texttt{num\_layers=9}, \texttt{lang="en"},
\texttt{idf=False}, and \texttt{rescale\_with\_baseline=False}. We
use batch size 16 for this metric. For SimCSE cosine similarity, we use
\texttt{sup-simcse-roberta-large} with the Hugging Face
\texttt{AutoTokenizer} and \texttt{AutoModel}; texts are tokenized with
padding and truncation to \texttt{max\_length=128}. We take the
\texttt{pooler\_output}, apply $\ell_2$ normalization, and compute the
cosine similarity between the reference and generated-answer
embeddings. The SimCSE batch size is 64.

\section{Extended Experiments}

\subsection{PCA Visualization of ChatTime~\cite{chattime}'s TS Embeddings}

To preliminarily validate our hypothesis that preserving the continuity and ordinality of TS Tokens in the embedding space is essential for effective time series modeling, we examine ChatTime, a representative token-based TS-LLM, by visualizing the geometry of its TS embeddings, which incorporates a vocabulary of 20,001 additional TS tokens.

\begin{figure}[h] 
    \centering
    \begin{subfigure}[b]{0.48\linewidth}
        \centering
        \includegraphics[width=\linewidth]{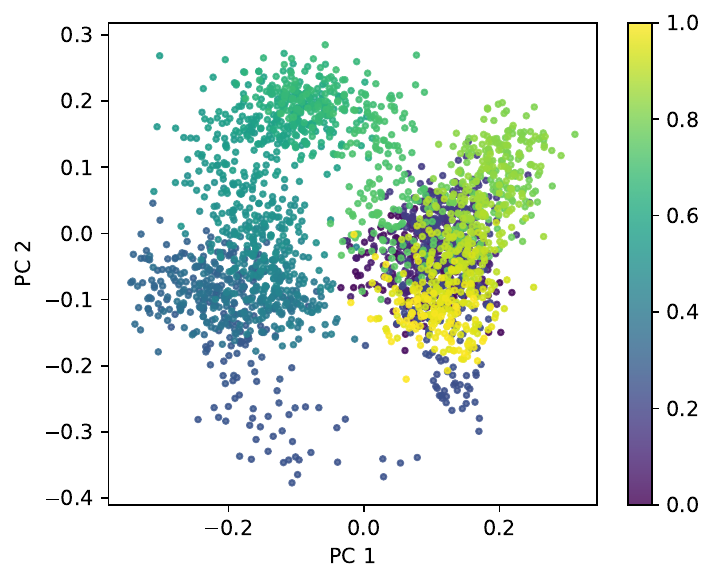}
        \caption{PCA-based 2D Visualization of TS Embeddings}
        \label{fig:vis_2d}
    \end{subfigure}
    \begin{subfigure}[b]{0.48\linewidth}
        \centering
        \includegraphics[width=\linewidth]{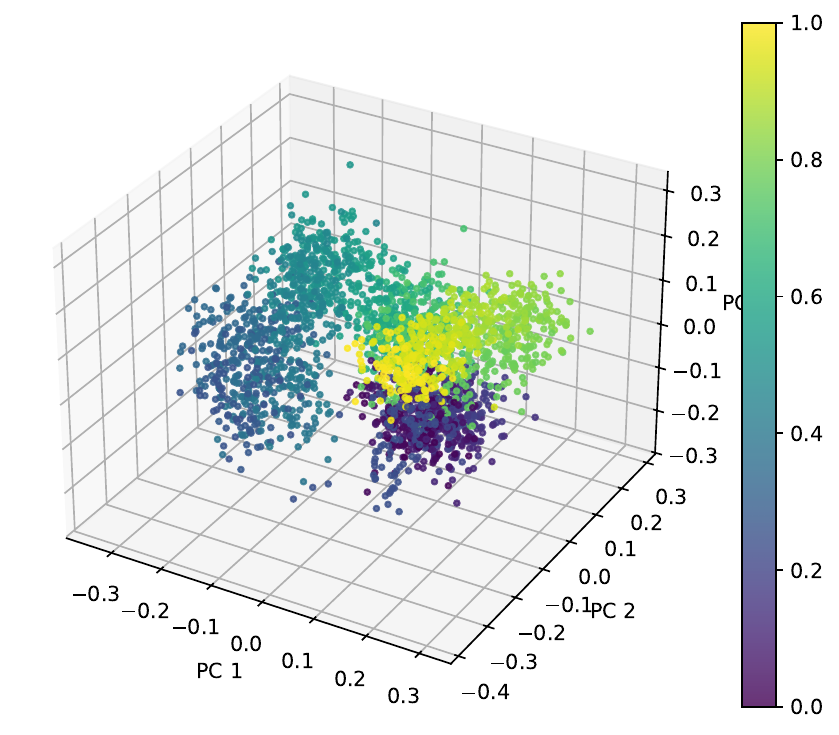}
        \caption{PCA-based 3D Visualization of TS Embeddings}
        \label{fig:vis_3d}
    \end{subfigure}
    
    \caption{The PCA visualization of TS-Token embeddings in ChatTime. The color from blue to yellow represents time series values ranging from $-1$ to $+1$. Notably, the TS tokens exhibit emerging continuous and ordered structures.}
    \label{fig:chattime_visualization}
\end{figure}

As illustrated in the PCA-based 2D and 3D visualizations (Figure~\ref{fig:chattime_visualization}), interestingly, although ChatTime treats these TS tokens merely as discrete words from a general vocabulary without explicit constraints, the learned embeddings exhibit emerging continuous and ordered structures. This demonstrates that TS tokens spontaneously evolve towards meaningful continuity and ordinality through large-scale training.

\subsection{Performance on MTMS-InWild Dataset}

To examine whether the proposed COM strategy generalizes to more sophisticated time series reasoning scenarios, we further evaluate model variants with different initialization schemes on the \textbf{MMTS-InWild} dataset. MMTS-InWild consists of two major task categories, \textit{analysis} and \textit{reasoning}, covering a broad range of subtasks that reflect the challenges of complex real-world time series understanding.

As shown in Table~\ref{tab:inwild_init}, initialization schemes that better preserve continuity and ordinality consistently outperform variants lacking such geometric structures. These results demonstrate the effectiveness and robustness of hard constraints, suggesting that continuity- and ordinality-aware initialization can benefit not only basic time series analysis but also complex time series reasoning.

\begin{table*}[ht]
\centering
\caption{Performance on the MMTS-InWild different across different initialization schemes. 
Analysis tasks include basic, noise, seasonality, trend, and volatility analysis, whereas reasoning tasks include analogical, causal, counterfactual, deductive, and inductive reasoning. 
The baseline corresponds to \texttt{Qwen2.5-3B-Instruct} with plain-text time series input.}
\label{tab:inwild_init}
\resizebox{0.85\textwidth}{!}{
\begin{tabular}{@{}l|cccccccc|c@{}}
\toprule
\textbf{Variant} & \textbf{Default} & \textbf{Slerp} & \textbf{PCA-Main} & \textbf{Helix 1} & \textbf{Helix 2} & \textbf{Helix 4} & \textbf{Helix 8} & \textbf{VMF} & \textbf{Baseline} \\ 
\midrule
Analysis  & 0.3619 & 0.4176 & \textbf{0.4354} & 0.4298 & 0.4031 & 0.4198 & 0.4053 & 0.3909 & 0.3742 \\
Reasoning & 0.3670 & 0.4063 & \textbf{0.4312} & 0.4102 & 0.3997 & 0.4194 & 0.4076 & 0.3853 & 0.3617 \\ 
\midrule
Total     & 0.3635 & 0.4068 & \textbf{0.4271} & 0.4253 & 0.3994 & 0.4170 & 0.3994 & 0.3828 & 0.3708 \\ 
\bottomrule
\end{tabular}
}
\end{table*}

\subsection{Efficacy of Self-Distillation Alignment}
\label{app:efficacy_self_alignment}

As introduced in Section~\ref{subsec:training}, our full training pipeline consists of two phases: \textit{self-distillation alignment} and \textit{downstream fine-tuning}. To examine the effectiveness of the self-distillation alignment phase, we adopt Slerp initialization and evaluate the model on the TSQA~\cite{chattime} dataset, comparing the \textit{Aligned} and \textit{Unaligned} configurations.

\begin{table}[h!]
\centering
\caption{Accuracy (\%) comparison of models with and without the \textit{self-distillation alignment} phase.}
\label{tab:efficacy_self_alignment}
\begin{small}
\begin{sc}
\begin{tabular}{lcccc}
\toprule
\multirow{2.5}{*}{Setting} & \multicolumn{2}{c}{\textbf{Unaligned}} & \multicolumn{2}{c}{\textbf{Aligned}} \\
\cmidrule(r){2-3} \cmidrule(l){4-5}
& \textbf{SID} & \textbf{ID} & \textbf{SID} & \textbf{ID} \\
\midrule
\textbf{Trend}       & 96.71 & 99.67 & 98.27 & 99.67 \\
\textbf{Seasonality} & 68.98 & 71.24 & 73.49 & 74.75  \\
\textbf{Volatility}  & 94.50 & 99.03 & 95.07 & 99.43  \\
\textbf{Outliers}    & 95.48 & 99.74 & 98.78 & 99.83  \\
\midrule
\textbf{Avg.}       & 88.94 & {92.44} & 91.40 & {93.52} \\
\bottomrule
\end{tabular}
\end{sc}
\end{small}
\end{table}

We consider two evaluation settings: \textbf{ID} (in-distribution) and \textbf{SID} (semi-in-distribution). In the ID setting, the training data are drawn from the same distribution as the TSQA test set. In the SID setting, we instead use fine-tuning data from ChatTime~\cite{chattime}; although the query format is consistent with that of TSQA, the underlying time series are from different distributions.

As shown in Table~\ref{tab:efficacy_self_alignment}, incorporating the self-distillation alignment phase consistently improves performance over the unaligned counterparts under both ID and SID settings. The gains appear not only in overall accuracy but also across all individual subtasks, demonstrating the robustness and effectiveness of self-distillation alignment.


\subsection{Performance on CGTSF-PTF Dataset}

To assess the broader applicability of our model to time series forecasting, we evaluate it on the CGTSF-PTF dataset~\cite{chattime}. 
The task requires predicting the next 24 time steps given historical observation windows of lengths $\{48, 72, 96, 120\}$; detailed dataset information is provided in Appendix~\ref{app:detail_dataset}. 
To ensure a fair comparison, all baselines follow the same normalization procedure as our model, as detailed in Section~\ref{sec:architecture}.

\begin{figure}[htbp]
    \centering

    \begin{subfigure}[b]{\linewidth}
        \centering
        \includegraphics[width=\linewidth]{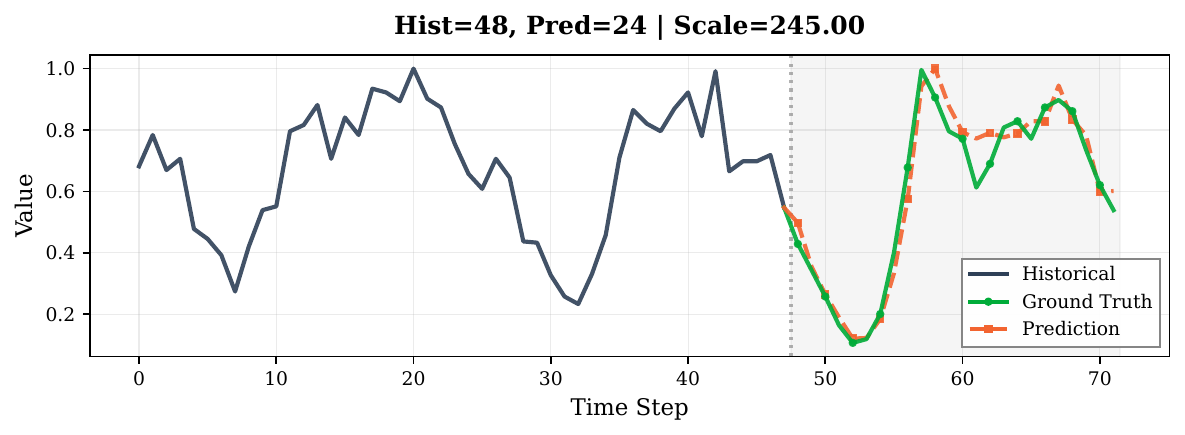}
    \end{subfigure}
    
    \begin{subfigure}[b]{\linewidth}
        \centering
        \includegraphics[width=\linewidth]{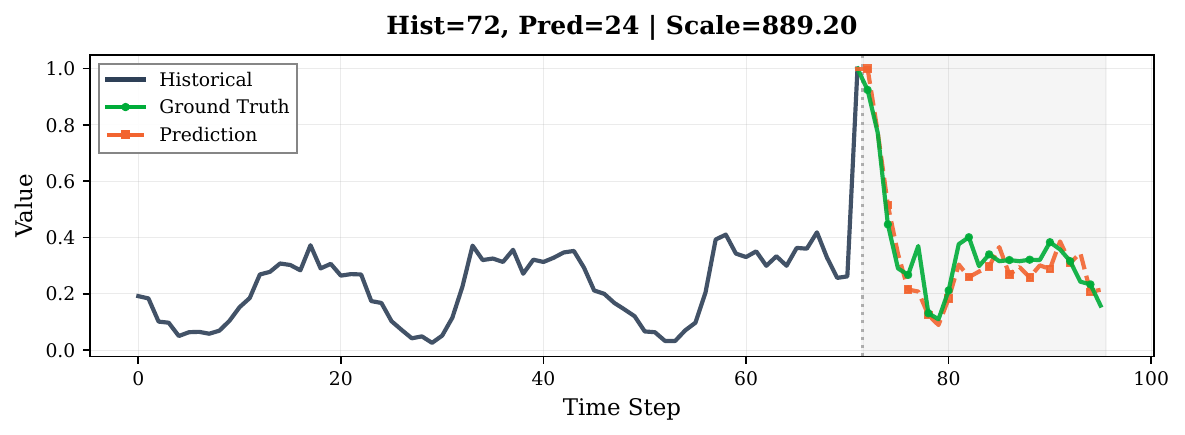}
    \end{subfigure}
    
    \begin{subfigure}[b]{\linewidth}
        \centering
        \includegraphics[width=\linewidth]{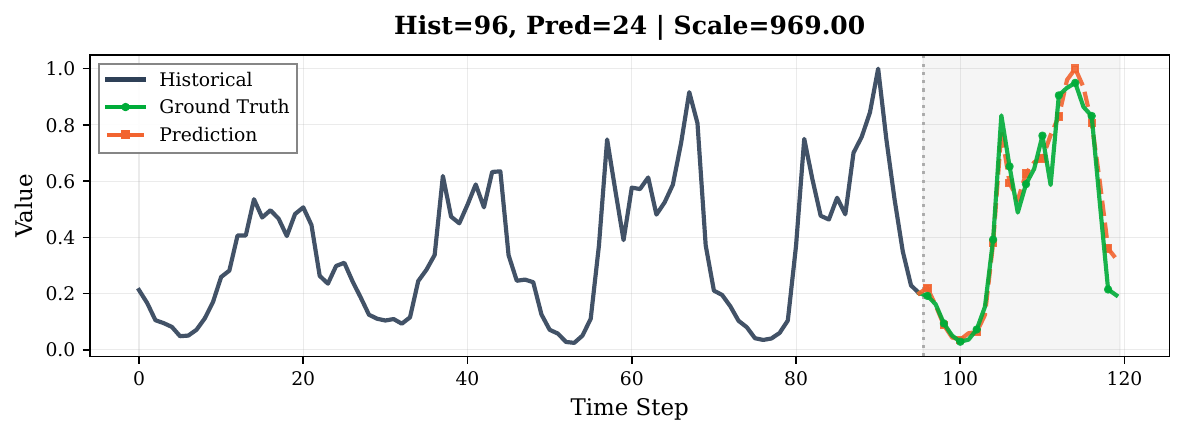}
    \end{subfigure}
    
    \begin{subfigure}[b]{\linewidth}
        \centering
        \includegraphics[width=\linewidth]{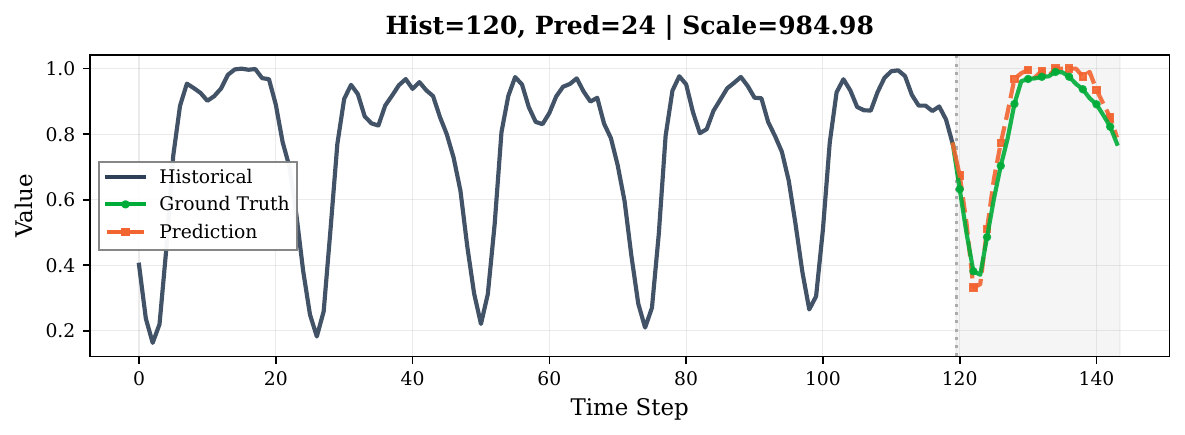}
    \end{subfigure}
    
    \caption{Visualization of our model's forecasting results on the CGTSF-PTF dataset. The four subplots correspond to input history lengths of $H \in \{48, 72, 96, 120\}$, respectively. These results demonstrate that our model achieves good forecasting performance and maintains robustness across varying history lengths and time series patterns.}
    \label{fig:forecast_plot}
\end{figure}

The quantitative results are reported in Table~\ref{tab:ptf_forecasting_hist_len}.
We implement all forecasting baselines based on the Time-Series-Library codebase~\cite{wang2026deep-tslib}, using its official implementations and recommended hyperparameter settings. For fair comparison, all baselines follow the same data preprocessing, normalization, and evaluation protocol as our method.
Although COM is not specifically designed as a dedicated forecasting architecture, it achieves highly competitive performance on CGTSF-PTF, obtaining the best or second-best results across most settings. 
In particular, COM outperforms several established forecasting baselines under shorter historical windows and remains comparable to strong specialized models under longer histories. 
These results suggest that, when equipped with appropriate time series token embeddings, token-based TS-LLMs can serve as an effective and competitive formulation for time series forecasting. 
We further visualize representative results in Figure~\ref{fig:forecast_plot}, showing that our model can capture major temporal variations across different history lengths and time series patterns.

\begin{table*}[htbp]
\centering
\caption{Forecasting performance on the PTF dataset under different history lengths. MSE and MAE are reported as evaluation metrics, where lower values indicate better performance. The best and second-best results are highlighted in bold and underlined, respectively.}
\label{tab:ptf_forecasting_hist_len}
\small
\setlength{\tabcolsep}{4pt}
\renewcommand{\arraystretch}{1.08}
\begin{tabular}{@{}l|cccccccc@{}}
\toprule
\multirow{3}{*}{\textbf{Model}} 
& \multicolumn{8}{c}{\textbf{History Length}} \\
\cmidrule(lr){2-9}
& \multicolumn{2}{c}{\textbf{48}} 
& \multicolumn{2}{c}{\textbf{72}} 
& \multicolumn{2}{c}{\textbf{96}} 
& \multicolumn{2}{c}{\textbf{120}} \\
\cmidrule(lr){2-3} \cmidrule(lr){4-5} \cmidrule(lr){6-7} \cmidrule(lr){8-9}
& \textbf{MSE} & \textbf{MAE} 
& \textbf{MSE} & \textbf{MAE} 
& \textbf{MSE} & \textbf{MAE} 
& \textbf{MSE} & \textbf{MAE} \\
\midrule
TimesNet~\cite{wu2022timesnet}        
& \underline{0.0213} & \underline{0.0928} 
& \underline{0.0152} & \textbf{0.0749} 
& \textbf{0.0113} & \textbf{0.0649} 
& \textbf{0.0107} & \textbf{0.0636} \\

iTransformer~\cite{liu2024itransformer}   
& 0.0321 & 0.1118 
& 0.0232 & 0.0947 
& 0.0186 & 0.0858 
& 0.0171 & 0.0811 \\

TimeXer~\cite{wang2024timexer}         
& 0.0403 & 0.1246 
& 0.0351 & 0.1178 
& 0.0239 & 0.0972 
& 0.0242 & 0.0987 \\

PatchTST~\cite{Yuqietal-2023-PatchTST}        
& 0.0438 & 0.1323 
& 0.0337 & 0.1139 
& 0.0247 & 0.0985 
& 0.0218 & 0.0910 \\

TimeMixer~\cite{wang2024timemixer}       
& 0.0462 & 0.1337 
& 0.0359 & 0.1158 
& 0.0253 & 0.0986 
& 0.0226 & 0.0926 \\

DLinear~\cite{zeng2023transformers-dlinear}         
& 0.0722 & 0.1701 
& 0.0552 & 0.1487 
& 0.0418 & 0.1314 
& 0.0371 & 0.1236 \\

Autoformer~\cite{wu2021autoformer}      
& 0.0871 & 0.2041 
& 0.0629 & 0.1659 
& 0.0462 & 0.1484 
& 0.0434 & 0.1461 \\

\midrule
DeepSeek-V4-Pro~\cite{deepseekai2026deepseekv4} 
& 0.0443 & 0.1389 
& 0.0396 & 0.1293 
& 0.0356 & 0.1207 
& 0.0348 & 0.1224 \\

Gemma3-12B-it~\cite{gemmateam2025gemma3technicalreport}
& 0.0955 & 0.2177
& 0.0800 & 0.1990
& 0.0661 & 0.1792
& 0.0637 & 0.1752 \\

Qwen3-14B~\cite{yang2025qwen3}
& 0.0951 & 0.2248
& 0.1058 & 0.2399
& 0.0863 & 0.2193
& 0.0803 & 0.2109 \\

Qwen2.5-3B~\cite{qwen2025qwen25technicalreport-qwen2.5}
& 0.1063 & 0.2286
& 0.1096 & 0.2386
& 0.0894 & 0.2097
& 0.0990 & 0.2265 \\

Qwen2.5-14B~\cite{qwen2025qwen25technicalreport-qwen2.5}
& 0.1833 & 0.3408
& 0.1520 & 0.3173
& 0.1386 & 0.3002
& 0.1421 & 0.3040 \\

\midrule
COM (ours)-3B      
& \textbf{0.0179} & \textbf{0.0820} 
& \textbf{0.0143} & \underline{0.0755} 
& \underline{0.0120} & \underline{0.0703} 
& \underline{0.0121} & \underline{0.0721} \\
\bottomrule
\end{tabular}
\end{table*}


\section{Artifact Licenses and Terms of Use}
All datasets, pretrained models, and software packages used in this work
are publicly available research artifacts. We use them in accordance
with their original licenses and terms of use. Specifically, the
forecasting baselines are implemented based on the Time-Series-Library
codebase, while the language models, embedding models, and evaluation
models used in our experiments retain their original licenses and usage
restrictions. We do not redistribute third-party datasets or pretrained
model weights. For the code and artifacts released with this paper, we
will specify the corresponding license in the public repository.



\end{document}